\documentclass{article}
\PassOptionsToPackage{numbers,compress}{natbib}
\usepackage[main,final]{neurips_2026}

\usepackage[utf8]{inputenc} 
\usepackage[T1]{fontenc}    
\usepackage{booktabs}
\usepackage{multirow}
\usepackage{graphicx}
\usepackage{subcaption}
\usepackage{hyperref}
\usepackage{tabularx}
\usepackage{titletoc}
\usepackage{threeparttable}
\usepackage{url}
\usepackage[most]{tcolorbox} 
\usepackage{amsmath,amssymb,mathtools,bm,amsthm}

\usepackage[ruled,vlined,linesnumbered]{algorithm2e}
\SetKwInput{KwRequire}{Require}
\SetKwInput{KwOutput}{Output}

\usepackage{microtype}
\usepackage{makecell}
\theoremstyle{remark}
\usepackage{listings}
\usepackage{xcolor}
\usepackage{colortbl}

\usepackage{wrapfig}
\usepackage{siunitx}

\theoremstyle{plain}
\newtheorem{theorem}{Theorem}[section]

\newtheorem{corollary}[theorem]{Corollary}
\theoremstyle{definition}

\newtheorem{assumption}[theorem]{Assumption}
\theoremstyle{remark}

\title{Denoising Time Matters:\\Diverse Generation in Diffusion Language Models}

\author{%
Jingxuan Wu\textsuperscript{1,*} \quad
Zhenglin Wan\textsuperscript{2,*} \quad
Yuzhe Yang\textsuperscript{3} \quad
Yiqiao Huang\textsuperscript{4} \\
\bfseries
Chubin Zhang\textsuperscript{5} \quad
Xingrui Yu\textsuperscript{6,7,$\dagger$} \quad
Ivor Tsang\textsuperscript{5,6,7} \quad
Yang You\textsuperscript{2} \\[0.8em]
\normalfont\small
\textsuperscript{1}The University of North Carolina at Chapel Hill \\
\textsuperscript{2}National University of Singapore \\
\textsuperscript{3}University of California, Santa Barbara \\
\textsuperscript{4}Harvard University \\
\textsuperscript{5}Nanyang Technological University \\
\textsuperscript{6}CFAR, Agency for Science, Technology and Research, Singapore \\
\textsuperscript{7}IAIC, Agency for Science, Technology and Research, Singapore \\[0.4em]
\textsuperscript{*}Equal contribution \qquad
\textsuperscript{$\dagger$}Corresponding author:
\texttt{Yu\_Xingrui@a-star.edu.sg}%
}

\begin{document}

\maketitle

\begin{abstract}
Diffusion language models (Diffusion-LMs) generate text through iterative denoising, exposing a temporal structure that is largely absent from autoregressive decoding. In this paper, we show that this temporal structure provides a useful control axis for generation diversity: early denoising steps mainly determine high-level semantic trajectories, while later steps refine lexical realization. Motivated by this observation, we propose Time-Annealed Perturbation Sampling (TAPS), a training-free inference strategy that samples nearby conditioning trajectories through time-aware, manifold-constrained perturbations. TAPS encourages semantic branching during early denoising and anneals the perturbation away before refinement, improving exploration while preserving prompt alignment, generation quality, and reasoning ability. Experiments on multiple Diffusion-LM backbones, including non-autoregressive and semi-autoregressive models, show that TAPS consistently improves semantic and lexical diversity across open-ended and instruction-following generation tasks, while preserving reasoning ability on verifiable reasoning benchmarks with negligible overhead.
\end{abstract}

\section{Introduction}

Diffusion language models (Diffusion-LMs) have recently emerged as a promising alternative to autoregressive language models for natural language generation \citep{cheng2025sdar, nie2025large, wang2025revolutionizing, wu2025fast,ye2025dream}. By generating text through iterative denoising rather than next token prediction, Diffusion-LMs introduce a distinct inference dynamics that has shown strong performance across writing, reasoning, and instruction-following tasks \citep{li2025survey, tae2025tess}. However, despite this progress, their diversity behavior remains insufficiently understood. Similar to autoregressive models, Diffusion-LMs can still produce repetitive or conservative outputs under diversity-seeking prompts \citep{zhang2024forcing}. More importantly, existing diversity-control methods are largely designed for autoregressive decoding and do not directly exploit the denoising-time structure of Diffusion-LMs. This raises a question: how can we use the temporal structure of diffusion inference to improve output diversity without degrading generation quality?

\noindent Existing diversity-control methods mainly operate through either training-time optimization or inference-time decoding. Training-based approaches modify supervised fine-tuning or reinforcement learning objectives to encourage diverse outputs \citep{li2024preserving,lanchantin2025diverse,li2025jointly}, but require additional data, optimization, and task-specific design. Inference-time methods are more flexible, including prompt-based strategies that modify or augment the input context \citep{zhang2025noveltybench,ruan2025g2}, and sampling-based methods that reshape the token distribution through temperature scaling or nucleus sampling \citep{holtzman2019curious,peeperkorn2024temperature,zhu2024hot}. However, these methods are largely designed for autoregressive decoding, where diversity is controlled through fixed input changes or token-level sampling, and therefore do not explicitly exploit the iterative denoising structure of Diffusion-LMs. Recent evidence further suggests that input-side perturbations can be more effective for eliciting creativity than output-level randomness alone~\citep{nagarajan2025roll}, motivating us to study how such perturbations should be scheduled across diffusion time. To address this gap, we consider where exploration should be introduced during diffusion inference. A natural hypothesis is that diversity control should follow the model's uncertainty: perturbations are most useful when the generation trajectory remains uncertain and plastic, but should be reduced once the model commits to a stable output. In Section~\ref{sec:temporal_structure}, we empirically show that Diffusion-LMs exhibit such a temporal structure: early denoising steps are more influential for high-level semantic variation, whereas later steps mainly refine lexical realization. Motivated by this observation, we propose Time-Annealed Perturbation Sampling (TAPS)\footnote{Code: \url{https://github.com/Johnny221B/TAPS}}, which applies conditioning perturbations during early denoising and anneals them away before late-stage refinement, encouraging semantic branching while preserving prompt alignment and generation quality. This quality--diversity balance may also benefit group-based reinforcement learning, where nearly identical rollouts provide little within-group variation. We consider TAPS a potential training-free tool for rollout collection; its effect on RL training remains to be evaluated.

\noindent Our contributions are threefold. 
(i) We identify a temporal semantic structure in Diffusion-LMs: early denoising drives high-level semantic branching, while later steps refine lexical realization. 
(ii) We propose TAPS, a training-free inference method that applies time-annealed, manifold-constrained conditioning perturbations to improve diversity without sacrificing quality. 
(iii) We show that TAPS improves diversity across open-ended and instruction-following generation, and increases the structural diversity of correct solutions on reasoning and coding tasks while maintaining competitive accuracy across multiple diffusion-based backbones.

\section{Related Work}


Diversity is a fundamental property of conditional text generation, as a single prompt often admits multiple valid outputs across tasks such as dialogue, story generation, protein design, and creative ideation~\citep{li2016diversity,fan2018hierarchical,gruver2023protein,si2024llmsgeneratenovelresearch}. Beyond surface-level variation, diverse generations can improve candidate selection, iterative refinement, and exploration in self-training or feedback-driven inference loops~\citep{stiennon2020learning,corso2023particleguidancenoniiddiverse,sadat2023cads,wu2025oscarorthogonalstochasticcontrol,du2024exploration,chen2025seed,cheng2025reasoning,novikov2025alphaevolve, yu2025imitationdiversebehaviorswasserstein,zhang2026evolvingdiffusionflowmatching, pan2026savlaspatiallyawareflowmatchingvisionlanguageaction}. Existing approaches to improving diversity can be broadly grouped into training-free and training-based methods.

\noindent\textbf{Training-free Methods For Diversity.}
Training-free diversity methods mainly operate at decoding time by modifying token selection to navigate the quality--diversity trade-off. Deterministic methods such as greedy decoding and beam search tend to follow high-probability trajectories and often produce repetitive outputs, with beam search incurring additional inference overhead \citep{freitag2017beam}. Stochastic sampling methods, including temperature sampling \citep{ackley1985learning}, top-$k$ sampling \citep{fan2018hierarchical}, and top-$p$ sampling \citep{holtzman2019curious}, inject randomness by reshaping or truncating the token distribution, but aggressive sampling can degrade coherence and fluency. To better adapt to context-dependent uncertainty, later methods dynamically adjust the sampling distribution using entropy or confidence signals, such as $\eta$-sampling and mirostat \citep{basu2020mirostat,hewitt2022truncation}, entropy-driven temperature \citep{zhang2024edt}, KLD-based temperature adjustment \citep{chang2023kl}, contrastive decoding with adaptive plausibility constraints \citep{li2023contrastive}, and min-$p$ sampling \citep{nguyen2024turning}. Complementarily, prompt-based methods condition on prior generations or auxiliary instructions to steer later outputs, but may introduce semantic drift and additional inference cost \citep{ruan2025g2,zhang2024forcing}. Overall, these methods primarily control diversity through token-level distributions or static prompt conditioning, whereas our method targets the time-dependent conditioning trajectory of Diffusion-LMs.

\noindent\textbf{Training Methods For Diversity.}
Training-based methods improve diversity by modifying learning objectives or post-training signals. In supervised learning, prior work encourages diverse outputs through maximum mutual information, unlikelihood training, and target-shaping regularization \citep{li2016diversity,welleck2019neural,li2020data}, with recent studies further addressing diversity collapse during instruction tuning \citep{li2024preserving,mai2024improving}. Another line incorporates diversity into preference optimization or reinforcement learning, including diversity-aware preference pairs, entropy-regularized preference learning, creativity-oriented objectives, and online RL signals that reward rare but high-quality trajectories \citep{lanchantin2025diverse,slocum2025diverse,chung2025modifying,ismayilzada2025creative,he2024olympiadbench,li2025jointly, wan2025diversifyingpolicybehaviorsextrinsic}. While effective, these methods require additional training data, optimization, or reward modeling, whereas our approach is training-free and operates directly at inference time.

\section{Methodology}
\label{sec:method}

\subsection{Preliminary}

\noindent\textbf{Autoregressive Language Models.}
Autoregressive language models generate text by factorizing the conditional distribution as
$p(y \mid x) = \prod_{t=1}^{T} p(y_t \mid x, y_{<t}),$
and sample tokens sequentially from this distribution. Randomness is introduced at the token
level through stochastic decoding strategies such as temperature scaling or nucleus sampling,
leading to diversity across different decoding trajectories.

\noindent\textbf{Diffusion Language Models.} Diffusion-based language models generate text through an iterative denoising process that progressively refines a corrupted or masked sequence. Randomness in Diffusion-LMs arises not only from token sampling but also from the inference dynamics themselves, such as stochastic denoising updates and remasking decisions across steps \citep{li2022diffusion}.
Existing Diffusion-LMs can be broadly categorized into non-autoregressive and semi-autoregressive variants.
Non-autoregressive Diffusion-LMs update all tokens in parallel at each denoising step by predicting masked tokens and selectively remasking them based on model confidence. In contrast, semi-autoregressive DLMs generate text in a block-wise manner: blocks are produced sequentially, while tokens within each block are refined using a non-autoregressive diffusion process.

\subsection{Temporal Semantic Structure in Diffusion-LMs}
\label{sec:temporal_structure}

A key question for principled diversity control in Diffusion-LMs is: \emph{when should exploration be introduced during denoising?} We answer this question from two complementary perspectives: intrinsic denoising dynamics and perturbation sensitivity.

\begin{figure}[!h]
  \centering
  \includegraphics[width=\linewidth]{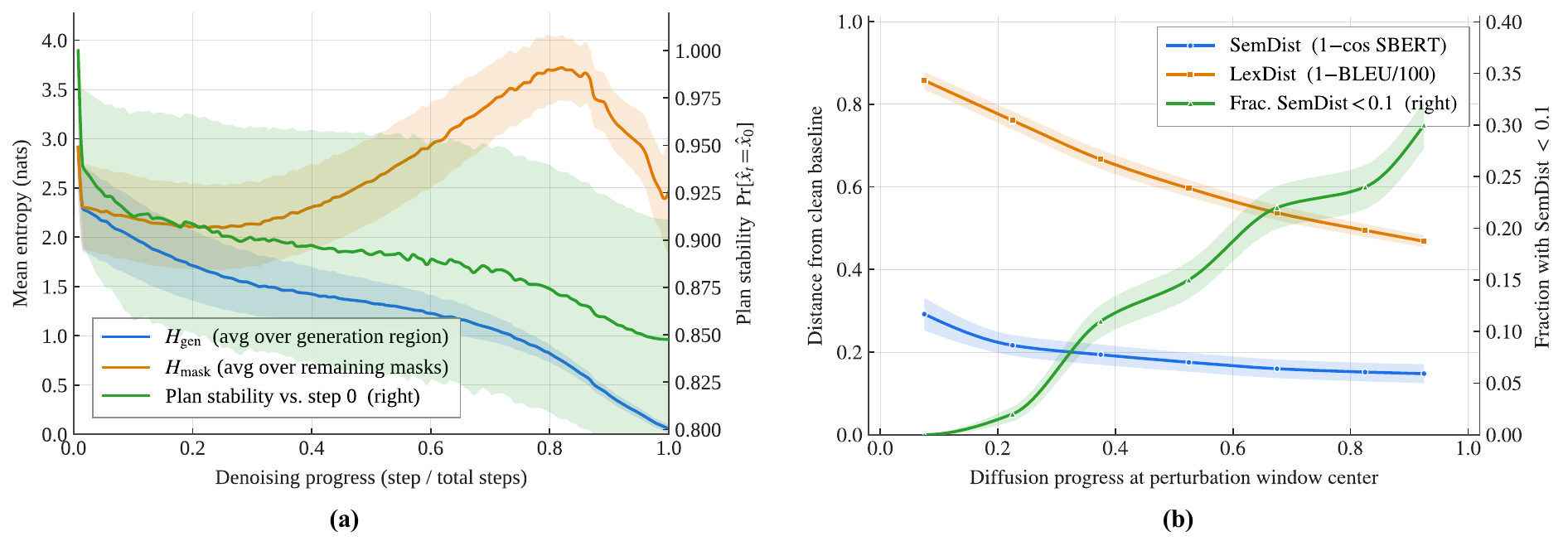}
  \caption{\small
    Evidence for temporal semantic structure in Diffusion-LMs.
    (a) Intrinsic denoising dynamics show that generation entropy decreases over time and the predicted plan stabilizes early, indicating an early high-plasticity phase followed by local refinement.
    (b) Windowed perturbation analysis shows that early conditioning perturbations induce larger semantic deviations, while later perturbations mainly affect lexical realization.
    Together, these results motivate TAPS to introduce exploration early and remove perturbations at a late stage.
  }
  \label{fig:temporal_structure}
\end{figure}

First, we examine the intrinsic uncertainty dynamics of Diffusion-LMs. As shown in Figure~\ref{fig:temporal_structure}(a), the model becomes increasingly certain about the generated content as denoising proceeds: generation entropy steadily decreases, and the predicted output plan quickly becomes close to the final generation. At the same time, the entropy over the remaining masked positions is non-monotonic, suggesting an easy-first decoding behavior where confident positions are resolved early and later steps focus on harder local refinements. This indicates that early stages are more uncertain, whereas later stages are increasingly committed to refinement. Second, we test whether this early plasticity has a stronger semantic impact through a windowed perturbation probe on 100 prompts. We inject Gaussian noise into the prompt embedding within a localized denoising window, slide it across the trajectory, and compare generations against clean baselines using SemDist ($1-\cos$ between Sentence-BERT embeddings~\citep{kirk2023understanding}) and LexDist ($1-\mathrm{BLEU}/100$~\citep{papineni2002bleu}). As shown in Figure~\ref{fig:temporal_structure}(b), early perturbations cause substantially larger semantic deviations, while later perturbations mainly affect lexical realization. Together, these results reveal a temporal division of labor in Diffusion-LMs, motivating TAPS to explore early and anneal perturbations away before late-stage refinement.

\subsection{Time-Annealed Perturbation Sampling}

The temporal semantic structure identified above reveals that multiple samples drawn from a standard Diffusion-LM share the same context embedding $\mathbf{E}$ throughout denoising, causing their early-stage semantic trajectories to converge and limiting output diversity. TAPS addresses this by introducing stochastic perturbations to the conditioning signal during the semantic formation phase, effectively allowing each sample to explore a different region of the conditional landscape.

Standard Diffusion-LM inference draws samples from $p(y \mid \mathbf{E})$, where diversity arises solely from the stochasticity of the denoising process. TAPS instead samples from a broader distribution induced by perturbing the conditioning signal:
\begin{equation}
\label{eq:taps_marginal}
p_{\text{TAPS}}(y) = \mathbb{E}_{\boldsymbol{\epsilon}}\left[p(y \mid \mathbf{E} + \sigma \boldsymbol{\epsilon})\right] = \int p(y \mid \tilde{\mathbf{E}}) \, \mathcal{N}(\tilde{\mathbf{E}} \mid \mathbf{E}, \sigma^2 \mathbf{I}) \, d\tilde{\mathbf{E}},
\end{equation}
which corresponds to a \emph{kernel-smoothed marginal} over a local neighborhood of the original conditioning. This formulation makes explicit why TAPS increases diversity: rather than sampling repeatedly from a single conditional $p(y \mid \mathbf{E})$, the model implicitly draws from a mixture of neighboring conditionals, encouraging exploration of distinct semantic modes. The perturbation scale $\sigma$ controls the radius of this neighborhood, where larger $\sigma$ induces broader exploration at the cost of deviating further from the intended conditioning.


\begin{figure}[!t]
    \centering
    \includegraphics[width=0.95\linewidth]{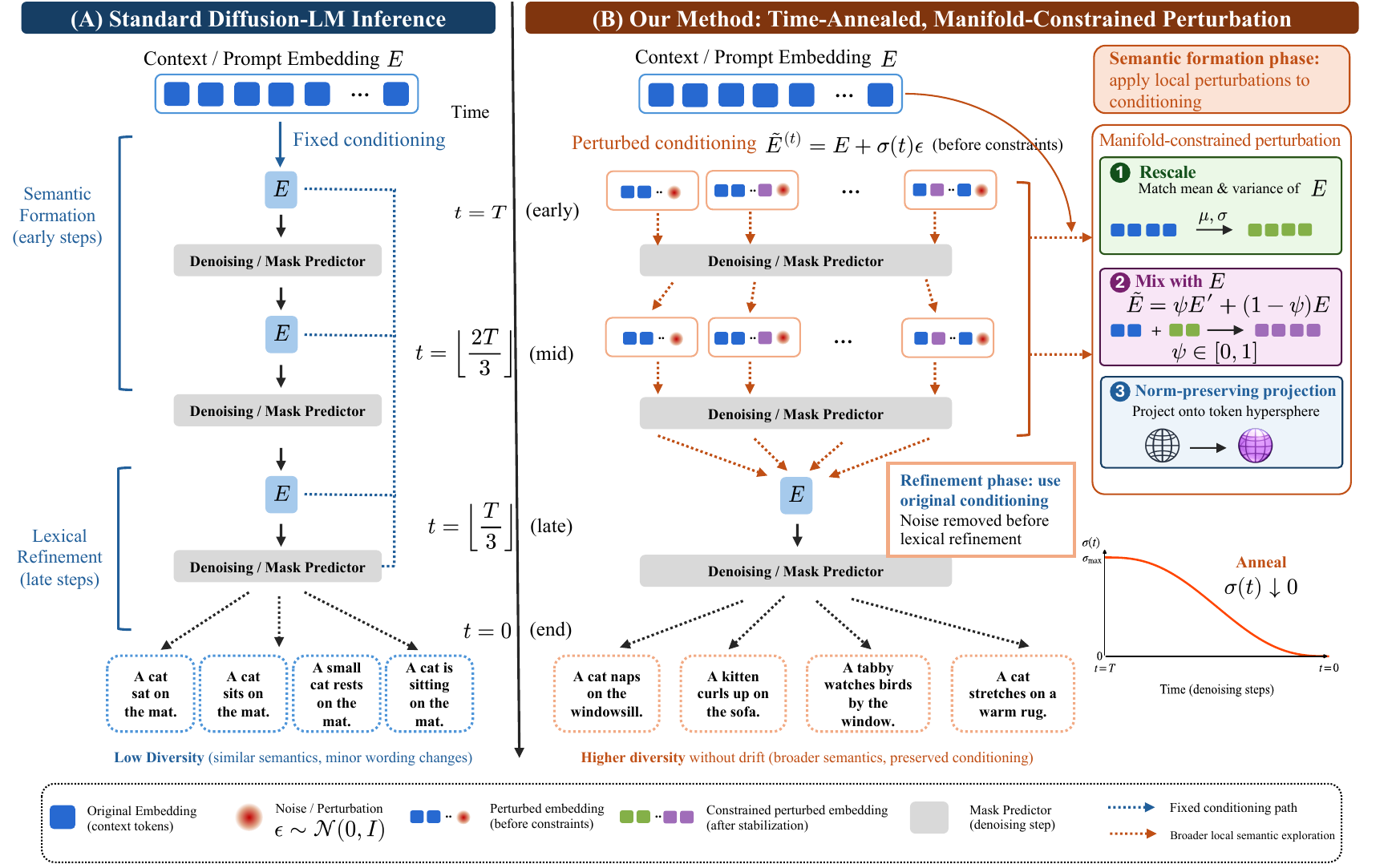}
    \caption{\small A conceptual comparison of the inference process between the base Diffusion-LM and TAPS, illustrating different context conditioning behaviors across denoising steps.}
    \label{fig:method}
\end{figure}

\textbf{Time-Annealed Perturbation.}
In practice, applying a fixed perturbation throughout the full denoising trajectory would corrupt the refinement phase and degrade generation quality. Guided by the temporal semantic structure, we anneal the perturbation strength over time: strong perturbations during the semantic formation phase encourage semantic branching, while the perturbation is progressively reduced and completely removed before the refinement phase to preserve lexical coherence and reasoning fidelity. As shown in Figure~\ref{fig:method}, we contrast the inference trajectories of TAPS and the base model, highlighting how time-annealed perturbations influence generation across denoising steps.

\noindent Formally, given a context embedding $\mathbf{E}$, the perturbed conditioning at inference step $t$ is defined as:
\begin{equation}
\label{eq:taps_perturb}
\tilde{\mathbf{E}}^{(t)} =
\begin{cases}
\mathbf{E} + \sigma(t)\boldsymbol{\epsilon}, & t \in [t_{\text{start}}, t_{\text{end}}], \\
\mathbf{E}, & \text{otherwise},
\end{cases}
\end{equation}
where $\boldsymbol{\epsilon} \sim \mathcal{N}(0, \mathbf{I})$ and $\sigma(t)$ follows a monotonically decreasing annealing schedule aligned with the transition from semantic formation to lexical refinement. Outside the active window $[t_{\text{start}}, t_{\text{end}}]$, the model conditions on the exact original embedding, ensuring that the refinement phase operates identically to standard inference. Although we focus on embedding-level perturbations in the main method, the same principle can also be instantiated at the token level by randomly masking context tokens to weaken conditioning strength. We study this token-level variant in Appendix~\ref{sec:token_noise}.


\subsection{Manifold-Constrained Perturbation}

The kernel-smoothed marginal in Eq.~\eqref{eq:taps_marginal} assumes that perturbed conditioning signals $\tilde{\mathbf{E}}$ remain within the valid region of the embedding space. Pretrained language-model representations have been shown to exhibit structured and anisotropic geometry rather than arbitrary isotropic variation~\citep{li2020sentence}. Naively adding isotropic Gaussian noise may therefore move the conditioning signal away from the region where the model was trained to operate, risking unstable or degraded generation. We therefore constrain the perturbed embeddings using three lightweight operations: distribution-aware rescaling, mixing with the original condition, and norm-preserving projection.

\textbf{Distribution-Aware Rescaling and Mixing.} The statistics-restoring rescaling follows the condition-annealing approach of CADS~\citep{sadat2023cads}. Unlike CADS, which interpolates between rescaled and unrescaled \emph{noised} conditions, TAPS mixes the rescaled embedding with the clean prompt embedding $\mathbf{E}$. Consequently, $\psi\to 0$ recovers the original conditioning. TAPS additionally applies token-wise norm-preserving projection to the resulting language-model embeddings. Let $\tilde{\mathbf{E}}^{(t)}$ denote the perturbed conditioning in Eq.~\eqref{eq:taps_perturb}. We first match its mean and standard deviation to those of the original embedding:
\[
\mathbf{E}' =
\frac{\tilde{\mathbf{E}}^{(t)}-\mu(\tilde{\mathbf{E}}^{(t)})}
{\sigma(\tilde{\mathbf{E}}^{(t)})}
\cdot \sigma(\mathbf{E})+\mu(\mathbf{E}),
\]
where $\mu(\cdot)$ and $\sigma(\cdot)$ are computed over token and embedding dimensions. We then interpolate with the original embedding:
\[
\hat{\mathbf{E}}=\psi \mathbf{E}'+(1-\psi)\mathbf{E},
\]
where $\psi\in[0,1]$ controls the effective perturbation radius. This keeps the perturbed condition close to the original prompt while allowing controlled exploration.

\textbf{Norm-Preserving Projection.}
To preserve token-level embedding geometry, we project each perturbed token embedding onto the hypersphere defined by the original token norm:
\[
\mathbf{E}^{\ast}_{i} =
\frac{\hat{\mathbf{E}}_{i}}
{\|\hat{\mathbf{E}}_{i}\|_2+\epsilon}
\cdot
\|\mathbf{E}_{i}\|_2,
\]
where $i$ indexes context tokens and $\epsilon$ is used for numerical stability. This retains the perturbation direction while preserving the original embedding magnitude.

Together, these operations constrain TAPS to sample nearby conditioning trajectories rather than arbitrary off-manifold perturbations, enabling stochastic exploration while maintaining prompt fidelity and generation stability. The complete algorithmic procedure, including time-annealed conditioning perturbation, distribution-aware rescaling, mixing, norm-preserving projection, and confidence-based token updates, is provided in Appendix~\ref{sec:algo}.

\textbf{Bounded Deviation and Reasoning Fidelity.}
The manifold-constrained design also provides a natural way to control how far TAPS deviates from standard inference. Since the effective perturbation radius is governed by the mixing coefficient $\psi$ and the annealed scale $\sigma(t)$, the accumulated deviation remains bounded by a finite perturbation budget. Moreover, because perturbations are annealed to zero before the refinement phase, later denoising updates operate under the original conditioning signal. This allows TAPS to explore alternative semantic or reasoning trajectories while remaining close to the original inference behavior, helping preserve generation quality and reasoning fidelity. We provide a theoretical analysis of bounded trajectory deviation and distributional shift in Appendix~\ref{sec:theory_taps}.

\section{Experiments}
\label{sec:exp}

\subsection{Experimental Setup}
\label{sec:exp_setup}

\noindent\textbf{Datasets \& Benchmarks.}
We evaluate TAPS across three settings: open-ended generation, instruction following, and verifiable reasoning. For open-ended generation, we use WritingPrompts~\citep{fan2018hierarchical} and NoveltyBench~\citep{zhang2025noveltybench}. For instruction-following and preference-aligned writing, we use the creative-writing subset of Arena-Hard-Auto~\citep{li2024crowdsourced}. For reasoning, we evaluate on mathematical reasoning benchmarks GSM8K~\citep{cobbe2021gsm8k} and MATH500~\citep{hendrycks2021measuring}, the science QA benchmark GPQA-Diamond~\citep{rein2023gpqa}, and the code generation benchmark HumanEval+~\citep{liu2023your}.

\noindent\textbf{Backbones.} 
We conduct experiments on four instruction-tuned diffusion language model backbones: LLaDA-8B-Instruct \citep{nie2025large}, Dream-Instruct-7B \citep{ye2025dream}, SDAR-8B-Chat \citep{cheng2025sdar}, and TraDo-8B-Instruct \citep{wang2025revolutionizing}. 
We report the main results on LLaDA and TraDo in the main paper, and provide additional results on Dream and SDAR in Appendix~\ref{sec:robustness_backbone}. Backbone hyperparameter settings are listed in Appendix~\ref{sec:hyper_set}.

\noindent\textbf{Baselines.} We compare against five training-free baselines that encourage diversity through complementary mechanisms while keeping the backbone, prompts, and generation seed. At the decoding level, we include standard token-distribution truncation methods, top-$k$, top-$p$, and min-$p$ \citep{nguyen2024turning}, which control diversity by restricting the candidate set in different ways. To account for diffusion-specific decoding behavior, we additionally evaluate entropy-based dynamic temperature (EDT) \citep{zhang2024edt}, which modifies the denoising dynamics during generation. Finally, we include Diverse Prompt as an input-level baseline that perturbs the prompt to elicit varied outputs without changing the model itself \citep{zhang2025noveltybench}. Detailed settings are provided in Appendix~\ref{sec:baseline_implementation}.

\noindent\textbf{Evaluation Metrics.} We evaluate diversity with lexical, semantic, and embedding-level metrics: IntraDistinct \citep{li2016diversity} and Div-BLEU ($1-$Self-BLEU) \citep{zhu2018texygen} for surface lexical diversity, Sent-BERT \citep{kirk2023understanding} for semantic variation, and Expectation-Adjusted Distinct Ngrams (EAD) \citep{liu2022rethinking} for embedding-space diversity; details are provided in Appendix~\ref{sec:eval_detail}. For quality evaluation, we use benchmark-specific automatic judges, including Skywork-Reward-Gemma-2-27B-v0.2~\citep{liu2024skywork} and GPT-4o-2024-08-06~\citep{hurst2024gpt}. We additionally report inference-time computational overhead in Appendix~\ref{sec:cost}.


\subsection{Results}

\subsubsection{Diversity under Multi-Sample Generation}

\begin{wraptable}{r}{0.5\columnwidth}
  \vspace{-10pt}
  \centering
  \setlength{\tabcolsep}{1.5pt} 
  \renewcommand{\arraystretch}{1.02} 
  \footnotesize 
  \caption{Diversity metrics comparison across two backbones on Novelty-Bench.}
  \label{tab:div_novelty_temp1.0}

  \resizebox{\linewidth}{!}{%
  \begin{tabular}{lcccc}
    \toprule
    & \multicolumn{4}{c}{\textbf{Diversity metrics}} \\
    \cmidrule(lr){2-5}
    \textbf{Backbone / Method} &
    \textbf{IntraDistinct}~$\uparrow$ &
    \textbf{Div-BLEU}~$\uparrow$ &
    \textbf{Sent-BERT}~$\uparrow$ &
    \textbf{EAD}~$\uparrow$ \\
    \midrule
    \multicolumn{5}{l}{\textbf{LLaDA-8B-Instruct}}\\
    \midrule
    Base (Normal)          & 72.47 & 48.44 & 25.80 & 50.22 \\
    Top-$p$                & 74.77 & 42.40 & 24.30 & 51.09 \\
    Top-$k$                & 75.64 & 45.16 & 23.38 & 50.50 \\
    Min-$p$                & 76.08 & 44.39 & 22.30 & 54.48 \\
    Diverse Prompt         & 71.50 & 24.32 & 19.84 & 48.58 \\
    \rowcolor{black!6}
    \textbf{TAPS}   & \textbf{78.32} & \textbf{66.26} & \textbf{36.04} & \textbf{63.35} \\
    \midrule
    \multicolumn{5}{l}{\textbf{TraDo-8B-Instruct}}\\
    \midrule
    Base (Normal)          & 83.73 & 57.06 & 24.28 & 62.15 \\
    Top-$p$                & 82.67 & 53.08 & 22.44 & 60.94 \\
    Top-$k$                & 82.54 & 56.49 & 22.93 & 62.84 \\
    Min-$p$                & 81.86 & 50.06 & 20.69 & 58.56 \\
    Diverse Prompt         & 85.57 & 61.50 & 28.58 & 66.51 \\
    \rowcolor{black!6}
    \textbf{TAPS}   & \textbf{86.50} & \textbf{65.88} & \textbf{29.84} & \textbf{67.31} \\
    \bottomrule
  \end{tabular}%
  }
  \vspace{-10pt}
\end{wraptable}

NoveltyBench consists of two subsets: Curated, with 100 carefully filtered prompts designed to elicit diverse, valid responses, and WildChat, containing real-world open-domain user instructions. We use the full Curated subset and randomly sample 500 WildChat prompts for computational efficiency. For each prompt, we generate 10 independent samples under identical decoding settings and compute all diversity and quality metrics over these sample sets, following the standard NoveltyBench protocol. Additional results are provided in Appendix~\ref{sec:diff_temp}.

\begin{wrapfigure}{r}{0.62\columnwidth}
\vspace{-2em}
  \centering
  \includegraphics[width=\linewidth]{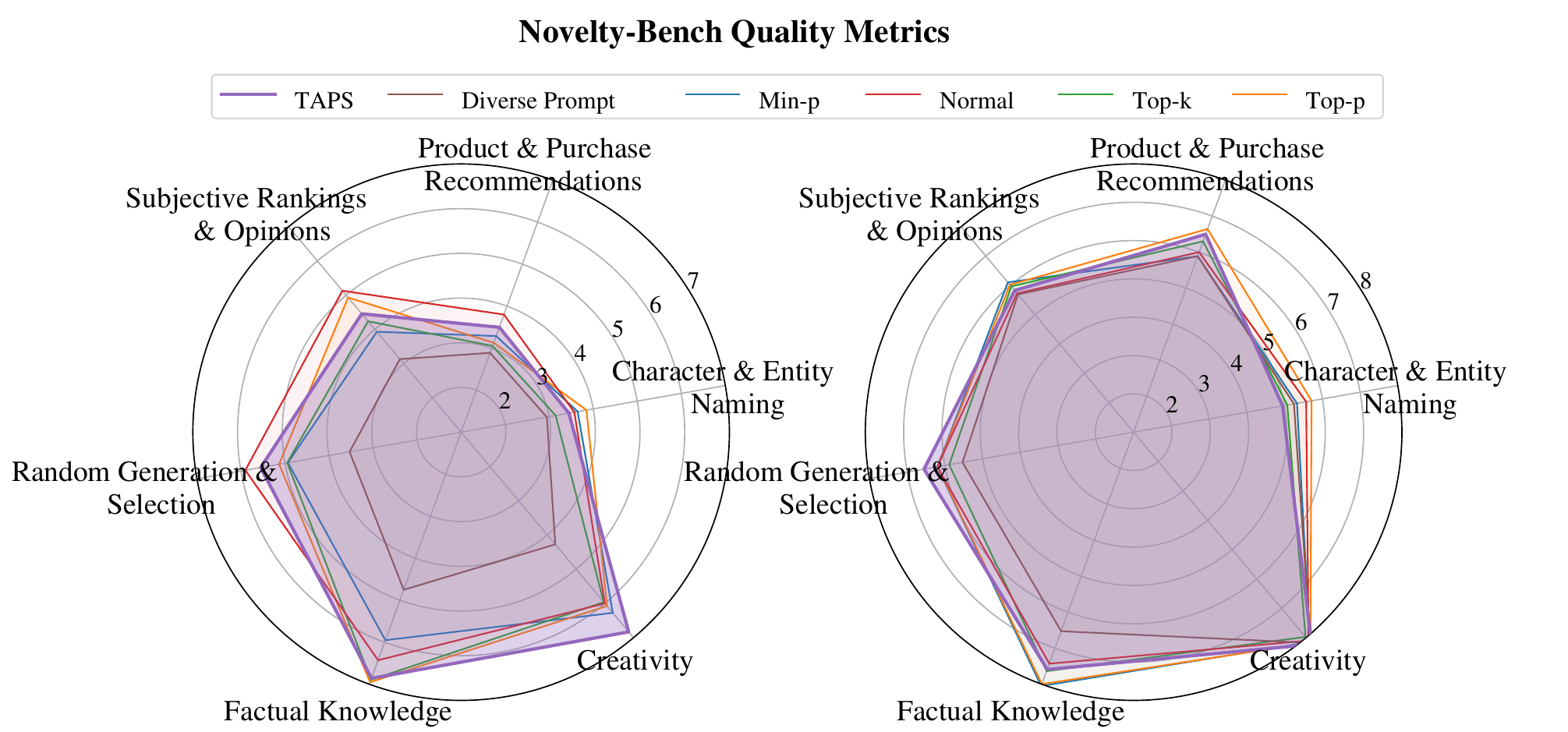}
  \caption{\small
    Quality comparison on NoveltyBench across preference dimensions. TAPS improves creativity-related scores while maintaining comparable overall quality; LLaDA and TraDo results are shown on the left and right, respectively.
    }
  \label{fig:radar_quality}
\vspace{-1em}
\end{wrapfigure}

Table~\ref{tab:div_novelty_temp1.0} reports diversity results aggregated over both subsets. Our method consistently achieves the best performance across all four diversity metrics for both LLaDA-8B-Instruct and TraDo-8B-Instruct, with especially large gains on Sent-BERT and EAD, indicating meaningful semantic and embedding-level variation beyond lexical changes. We further evaluate generation quality using Skywork-Reward-Gemma-2-27B-v0.2 across six human-preference dimensions (Figure~\ref{fig:radar_quality}). Our method performs strongest on creativity-related dimensions while remaining comparable to or better than standard decoding baselines on other aspects, showing that it improves diversity and creative quality without sacrificing usefulness or coherence.

\noindent We further observe that the effectiveness of several baselines varies markedly across backbones. While Diverse Prompt yields noticeable gains on LLaDA, its performance on TraDo deteriorates substantially across both diversity and quality metrics. A similar backbone-dependent trend is observed for min-$p$ decoding, which performs favorably on LLaDA but degrades on TraDo. In addition, we find that EDT fails to produce reasonable generations on NoveltyBench under our evaluation setting; despite extensive parameter tuning, EDT often leads to degenerate or incoherent outputs and is therefore omitted from the comparison. These observations suggest that prompt-based and token-level decoding strategies can be highly sensitive to the underlying model’s generative capacity and the task distribution. In contrast, our method exhibits more consistent improvements across backbones, indicating a more robust and model-agnostic mechanism for enhancing diversity.

\subsubsection{Open-Ended Story Generation}
\label{sec:open_ended_story}

Table~\ref{tab:div_quality_two_benchmarks_compact} presents results on WritingPrompts for long-form story generation. We randomly sample 250 prompts and generate 16 outputs per prompt. Under this evaluation protocol, our method consistently achieves the highest diversity scores across lexical, semantic, and embedding-level metrics on both backbones. In particular, we observe clear improvements in Sent-BERT and EAD, indicating that the induced diversity extends beyond surface-level lexical variation and reflects more substantial semantic differences among generated stories. Compared to token-level truncation baselines such as top-$k$, top-$p$, and min-$p$, our approach yields stronger and more balanced gains across all diversity metrics, suggesting more effective exploration of the narrative space.

\noindent To assess generation quality, we use GPT-4o as an automatic judge and report scores along four dimensions: creativity, coherence, writing quality, and relevance. As shown in Table~\ref{tab:div_quality_two_benchmarks_compact}, our method maintains quality scores that are comparable to, and in some cases slightly better than, baseline decoding strategies across all evaluated dimensions. Notably, the substantial gains in diversity do not lead to degraded coherence or relevance, indicating that the generated stories remain well-structured and faithful to the input prompts. In contrast, min-$p$ decoding exhibits consistent drops across diversity metrics, while EDT leads to noticeable degradation in quality scores, highlighting the difficulty of preserving narrative quality in long-form generation.

\begin{table}[!h]
\centering
\small
\setlength{\tabcolsep}{3.5pt}
\renewcommand{\arraystretch}{1.00}
\caption{Diversity and multi-aspect quality comparison across LLaDA and TraDo on two benchmarks.}

\resizebox{\linewidth}{!}{%
\begin{tabular}{lcccccccc}
\toprule
& \multicolumn{4}{c}{\textbf{Diversity metrics}} & \multicolumn{4}{c}{\textbf{Quality (GPT-4o)}} \\
\cmidrule(lr){2-5}\cmidrule(lr){6-9}
\textbf{Backbone / Method} &
\textbf{IntraDistinct}~$\uparrow$ &
\textbf{Div-BLEU}~$\uparrow$ &
\textbf{Sent-BERT}~$\uparrow$ &
\textbf{EAD}~$\uparrow$ &
\textbf{Crea}~$\uparrow$ &
\textbf{Coh}~$\uparrow$ &
\textbf{WQ}~$\uparrow$ &
\textbf{Rel}~$\uparrow$ \\
\midrule

\multicolumn{9}{l}{\textit{WritingPrompt}}\\[-1mm]
\addlinespace[0.5mm]
\midrule
\addlinespace[0.3mm]
\hspace{2mm}\textbf{LLaDA-8B-Instruct}        & 58.39 & 57.44 & 34.80 & 51.91 & 6.26 & 6.05 & \textbf{6.33} & 5.07 \\
\hspace{2mm}Top-$p$              & 58.51 & 57.16 & 34.76 & 51.96 & \textbf{6.34} & \textbf{6.07} & 6.31 & 4.93 \\
\hspace{2mm}Top-$k$              & 56.95 & 55.01 & 32.82 & 49.45 & 5.99 & 5.61 & 6.01 & 4.66 \\
\hspace{2mm}Min-$p$              & 55.90 & 53.86 & 32.20 & 47.62 & 6.18 & 5.92 & 6.27 & 4.74 \\
\hspace{2mm}Diverse Prompt                  & 57.01 & 42.40 & 22.50 & 42.15 & 6.16 & 5.42 & 5.91 & 4.37 \\
\hspace{2mm}EDT                  & 58.85 & 58.90 & 35.02 & 52.72 & 5.95 & 5.51 & 5.98 & 4.58 \\
\rowcolor{black!6}
\hspace{2mm}\textbf{TAPS} & \textbf{60.51} & \textbf{60.82} & \textbf{40.95} & \textbf{55.33}
                                 & 6.21  & 5.98  & 6.23  & \textbf{5.31} \\
\addlinespace[0.6mm]
\midrule
\addlinespace[0.3mm]
\hspace{2mm}\textbf{TraDo-8B-Instruct}        & 67.70 & 81.71 & 52.23 & 66.46 & 5.31 & 6.09 & 5.84 & 6.93 \\
\hspace{2mm}Top-$p$              & 61.61 & 78.50 & 53.54 & 59.76 & 5.28 & \textbf{6.21} & \textbf{5.92} & \textbf{7.05} \\
\hspace{2mm}Top-$k$              & 58.47 & 76.20 & 51.57 & 55.83 & 5.16 & 6.11 & 5.75 & 7.00 \\
\hspace{2mm}Min-$p$              & 55.54 & 75.08 & 51.37 & 51.80 & 5.06 & 6.11 & 5.78 & 6.97 \\
\hspace{2mm}Diverse Prompt                  & 61.64 & 77.36 & 39.06 & 65.35 & 5.23 & 6.17 & 5.87 & 6.94 \\
\hspace{2mm}EDT                  & 67.80 & 81.90 & 52.43 & 66.61 & 5.15 & 5.85 & 5.60 & 6.79 \\
\rowcolor{black!6}
\hspace{2mm}\textbf{TAPS} & \textbf{68.44} & \textbf{82.37} & \textbf{53.84} & \textbf{67.06}
                                 & \textbf{5.32}  & 6.10  & 5.83  & 6.96 \\

\addlinespace[1.2mm]
\midrule
\addlinespace[0.6mm]
\multicolumn{9}{l}{\textit{Arena-Hard-Auto (Creative-Writing)}}\\[-1mm]
\addlinespace[0.5mm]
\midrule
\addlinespace[0.3mm]
\hspace{2mm}\textbf{LLaDA-8B-Instruct}        & 54.02 & 67.79 & 24.80 & 44.72 & 4.49 & 4.46 & 4.48 & 5.07 \\
\hspace{2mm}Top-$p$              & 53.75 & 66.80 & 24.18 & 44.27 & 4.56 & 4.53 & 4.60 & \textbf{5.17} \\
\hspace{2mm}Top-$k$              & 52.80 & 63.48 & 22.78 & 43.28 & 4.44 & 4.51 & \textbf{4.61} & 5.08 \\
\hspace{2mm}Min-$p$              & 51.26 & 60.81 & 22.53 & 41.05 & 4.40 & 4.48 & 4.57 & 5.01 \\
\hspace{2mm}Diverse Prompt       & 50.48 & 40.51 & 17.72 & 29.64    & 4.47 & 4.43 & 4.45 & 4.90 \\
\hspace{2mm}EDT                  & 54.25 & 67.90 & 25.32 & 45.61 & 4.31 & 4.20 & 4.24 & 4.71 \\
\rowcolor{black!6}
\hspace{2mm}\textbf{TAPS} & \textbf{57.70} & \textbf{69.53} & \textbf{27.35} & \textbf{48.12}
                                 & \textbf{4.66}  & \textbf{4.59}  & 4.43  & 4.95 \\
\addlinespace[0.6mm]
\midrule
\addlinespace[0.3mm]
\hspace{2mm}\textbf{TraDo-8B-Instruct}        & 77.65 & 81.17 & 34.57 & 73.76 & \textbf{5.56} & 5.08 & 4.84 & 6.05 \\
\hspace{2mm}Top-$p$              & 76.90 & 77.81 & 32.36 & 70.78 & 5.52 & \textbf{5.39} & \textbf{5.16} & \textbf{6.17} \\
\hspace{2mm}Top-$k$              & 75.14 & 76.73 & 32.24 & 69.93 & 5.42 & 5.29 & 5.05 & 6.07 \\
\hspace{2mm}Min-$p$              & 73.25 & 74.41 & 29.90 & 65.55 & 5.47 & 5.22 & 5.14 & 6.16 \\
\hspace{2mm}Diverse Prompt       & 63.85 & 47.36 & 17.17 & 48.34 & 5.42 & 5.21 & 5.07 & 6.00 \\
\hspace{2mm}EDT                  & 78.20 & 81.92 & 34.80 & 73.91 & 5.20 & 4.33 & 4.27 & 5.56 \\
\rowcolor{black!6}
\hspace{2mm}\textbf{TAPS} & \textbf{80.90} & \textbf{82.93} & \textbf{35.65} & \textbf{74.66}
    & 5.39 & 5.24 & 5.08 & 6.12 \\
\bottomrule
\end{tabular}%
}

\label{tab:div_quality_two_benchmarks_compact}
\end{table}

\subsubsection{Preference-Aligned Creative Writing}

We further evaluate TAPS on the creative-writing subset of Arena-Hard-Auto, using 250 preference-aligned prompts from real user interactions and generating 8 outputs per prompt. As shown in Table~\ref{tab:div_quality_two_benchmarks_compact}, TAPS consistently achieves the highest diversity scores on both backbones, with especially clear gains on semantic and embedding-based metrics, indicating meaningful narrative variation beyond surface lexical changes. Using GPT-4o to assess four quality aspects following Section~\ref{sec:open_ended_story}, TAPS maintains quality comparable to strong baselines, while Diverse Prompt suffers a large diversity drop and EDT degrades quality. We also conduct pairwise preference evaluation with GPT-4o as the judge. The win-rate results in Table~\ref{tab:arena_hard_results} show that TAPS remains competitive with strong decoding baselines while offering substantially higher diversity, suggesting that its diversity gains do not introduce undesirable randomness. Additional evaluations are provided in the appendix, including preference results using GPT-4o and Gemini-2 (Appendix~\ref{sec:pref_model}), robustness results on SDAR and Dream (Appendix~\ref{sec:add_backbone_open}), and temperature sensitivity analysis (Appendix~\ref{sec:diff_temp}).

\subsubsection{Reasoning Robustness}
\label{sec:reason_exp}

We further evaluate reasoning robustness across four benchmarks: GSM8K, MATH500, GPQA-Diamond, and HumanEval+, which cover mathematical reasoning, scientific QA, and code generation. For all datasets, we use a unified decoding temperature of 0.7 and generate 5 independent samples per question under identical decoding conditions. We report three metrics: mean accuracy across samples, Pass@5, and majority-vote accuracy, comparing five decoding methods.

\begin{table}[!h]
\centering
\small
\setlength{\tabcolsep}{3.5pt}
\renewcommand{\arraystretch}{1.08}
\caption{Reasoning performance across four benchmarks. We report mean accuracy, majority-vote accuracy, and Pass@5. Mean accuracy is reported with 95\% confidence interval half-width.}
\label{tab:reasoning_results}
\resizebox{\linewidth}{!}{%
\begin{tabular}{lccc ccc ccc ccc}
\toprule
\multirow{2}{*}{\textbf{Method}} 
& \multicolumn{3}{c}{\textbf{GSM8K}}
& \multicolumn{3}{c}{\textbf{MATH-500}}
& \multicolumn{3}{c}{\textbf{GPQA-Diamond}}
& \multicolumn{3}{c}{\textbf{HumanEval+}} \\
\cmidrule(lr){2-4}\cmidrule(lr){5-7}\cmidrule(lr){8-10}\cmidrule(lr){11-13}
& Mean{\scriptsize$\pm$CI} & Maj. & Pass@5
& Mean{\scriptsize$\pm$CI} & Maj. & Pass@5
& Mean{\scriptsize$\pm$CI} & Maj. & Pass@5
& Mean{\scriptsize$\pm$CI} & Maj. & Pass@5 \\
\midrule
Base model 
& 75.7{\scriptsize$\pm$0.92} & 83.0 & 90.6 
& 17.5{\scriptsize$\pm$1.32} & 27.8 & 39.2 
& 17.1{\scriptsize$\pm$1.28} & 23.2 & 48.5 
& 39.6{\scriptsize$\pm$3.26} & 42.1 & 52.4 \\

Top-$p$    
& \textbf{77.2{\scriptsize$\pm$0.81}} & \textbf{84.5} & \underline{91.1} 
& 17.3{\scriptsize$\pm$1.30} & 28.2 & 39.4 
& 17.2{\scriptsize$\pm$1.28} & 25.2 & 47.0 
& \textbf{42.2{\scriptsize$\pm$3.65}} & \textbf{43.9} & 53.0 \\

Top-$k$    
& 75.9{\scriptsize$\pm$0.76} & 82.8 & 90.3 
& 17.6{\scriptsize$\pm$1.38} & 28.4 & 39.0 
& \textbf{19.2{\scriptsize$\pm$1.59}} & 25.8 & \underline{49.5} 
& 39.9{\scriptsize$\pm$2.96} & 42.1 & 53.0 \\

Min-$p$    
& 76.2{\scriptsize$\pm$0.83} & 83.6 & 89.8 
& \textbf{18.1{\scriptsize$\pm$1.34}} & \textbf{29.2} & \underline{42.0} 
& 18.5{\scriptsize$\pm$1.43} & \underline{28.3} & 48.5 
& 40.5{\scriptsize$\pm$2.96} & 42.7 & \underline{53.7} \\

\rowcolor{black!6}
\textbf{TAPS}       
& \underline{76.8{\scriptsize$\pm$0.88}} & \underline{84.2} & \textbf{91.2} 
& \underline{18.0{\scriptsize$\pm$1.24}} & \underline{28.8} & \textbf{42.2} 
& \underline{18.7{\scriptsize$\pm$1.03}} & \textbf{28.8} & \textbf{50.0} 
& \underline{40.8{\scriptsize$\pm$1.57}} & \underline{43.0} & \textbf{56.7} \\
\bottomrule
\end{tabular}%
}
\end{table}

As shown in Table~\ref{tab:reasoning_results}, TAPS maintains competitive mean accuracy while improving Pass@5 and majority-vote accuracy over the base model across all four benchmarks. These results suggest that the injected context-level perturbations do not simply introduce random noise: multi-sample inference remains effective even when individual generations explore different trajectories. Although exploration may slightly affect single-sample accuracy relative to some decoding baselines, aggregating multiple samples can still recover correct answers effectively. The results on mathematical reasoning, scientific QA, and code generation indicate that this behavior is not confined to a single type of task. Appendix~\ref{sec:gsm8k_examples} provides qualitative GSM8K examples comparing reasoning trajectories under different decoding strategies. We additionally evaluate reasoning performance on further Diffusion-LM backbones in Appendix~\ref{sec:add_backbone_reason}, showing that the reasoning results are not limited to the main backbone.

\noindent\textbf{Diversity of correct reasoning paths.}
The accuracy and aggregation results above do not establish whether correct samples follow genuinely different solutions rather than expressing similar solutions in different words. To examine this, we analyze 100 GSM8K questions and 60 HumanEval+ problems for which every compared method produces at least three correct outputs out of five. For each method, path-level diversity is computed only among its correct outputs on this shared set of tasks. Definitions and computation details for the metrics are provided in Appendix~\ref{sec:path_metric_details}.

\begin{table}[tbp]
\centering
\small
\caption{Diversity of correct solutions on GSM8K and HumanEval+.}
\label{tab:correct_path_diversity}
\resizebox{\linewidth}{!}{%
\begin{tabular}{lcccccc}
\toprule
& \multicolumn{3}{c}{\textbf{HumanEval+}} &
  \multicolumn{3}{c}{\textbf{GSM8K}} \\
\cmidrule(lr){2-4}\cmidrule(lr){5-7}
\textbf{Method}
& Norm-AST Diff. $\uparrow$
& Alg. Skeleton $\uparrow$
& Identical $\downarrow$
& Distinct-path $\uparrow$
& Op. Edit $\uparrow$
& Identical $\downarrow$ \\
\midrule
Base   & 41.72 & 17.78 & 58.28 & 31.60 & 19.75 & 6.33 \\
Top-$p$ & 35.78 & 11.17 & 64.22 & 31.27 & 19.87 & 6.60 \\
Top-$k$ & 42.94 & 19.22 & 57.06 & 26.33 & 18.04 & 6.80 \\
Min-$p$ & 33.44 & 11.06 & 66.56 & 26.53 & 16.28 & 6.60 \\
\rowcolor{black!6}
\textbf{TAPS} & \textbf{54.44} & \textbf{21.44} & \textbf{45.56}
              & \textbf{38.20} & \textbf{22.96} & \textbf{3.60} \\
\bottomrule
\end{tabular}%
}
\end{table}

As shown in Table~\ref{tab:correct_path_diversity}, TAPS produces more varied structures among correct HumanEval+ programs and more distinct correct reasoning paths on GSM8K than all compared decoding methods. It also yields the fewest identical solutions on both benchmarks. This supports the interpretation that TAPS encourages exploration of different valid solution paths, rather than merely changing surface wording. Together with the competitive accuracy, Pass@5, and majority-vote results in Table~\ref{tab:reasoning_results}, the path-level analysis shows that this additional exploration can be achieved without sacrificing overall multi-sample reasoning performance.

\subsection{Ablation Study}

\begin{wraptable}[9]{r}{0.58\columnwidth}
  \centering
  \footnotesize
  \setlength{\tabcolsep}{2.5pt}
  \renewcommand{\arraystretch}{1.06}
  \caption{\small Manifold perturbation ablation.}
  \label{tab:component_ablation}
  \resizebox{\linewidth}{!}{%
  \begin{tabular}{lccccc}
    \toprule
    \textbf{Variant} &
    \textbf{IntraDistinct} &
    \textbf{Div-BLEU} &
    \textbf{Sent-BERT} &
    \textbf{EAD} &
    \textbf{Quality} \\
    \midrule
    Raw perturbation & 65.82 & \textbf{57.25} & 34.49 & 53.55 & 6.73 \\
    w/o rescaling & 65.41 & 56.81 & 34.16 & 53.09 & 6.97 \\
    w/o mixing & \textbf{65.92} & 57.12 & 34.85 & \textbf{53.75} & 7.15 \\
    w/o norm projection & 65.17 & 57.08 & 34.93 & 53.15 & 7.11 \\
    \rowcolor{black!6}
    TAPS & 64.99 & 57.08 & \textbf{35.51} & 52.94 & \textbf{7.30} \\
    \bottomrule
  \end{tabular}%
  }
  \vspace{-1em}
\end{wraptable}

To examine whether the manifold-constrained perturbation design is necessary, we ablate each stabilization component in Table~\ref{tab:component_ablation}. 
Raw perturbation achieves the highest Div-BLEU but yields the lowest quality score, indicating that unconstrained embedding noise can destabilize generation. 
Removing rescaling, mixing, or norm-preserving projection also reduces quality, showing that matching embedding statistics, bounding the perturbation radius, and preserving token-wise geometry all contribute to stable generation. 
The full TAPS variant achieves the highest quality and the best semantic diversity measured by Sent-BERT, while maintaining competitive lexical and embedding-level diversity. 
These results suggest that manifold-constrained perturbations improve diversity through semantically meaningful variation rather than arbitrary off-manifold noise.

\subsection{Sensitivity Analysis}
This section explores the role of the most important parameters in our method on the final quality and diversity of generated samples. Additional analyses are provided in Appendix~\ref{sec:add_ablation}.

\begin{table*}[!h]
\centering
\caption{Sensitivity analysis of TAPS hyperparameters. Left: robustness under different noise scales and injection windows. Right: sensitivity to the mixing coefficient $\psi$.}
\label{tab:sensitivity_analysis}
\small
\setlength{\tabcolsep}{3pt}
\renewcommand{\arraystretch}{1.06}

\begin{subtable}[t]{0.49\textwidth}
\centering
\caption{Sensitivity to noise scale $\sigma$ and injection window $t$.}
\label{tab:robust_combined_singlecol}
\resizebox{\linewidth}{!}{%
\begin{tabular}{lcccc}
\toprule
\textbf{Backbone / Setting} &
\textbf{Div-BLEU}~$\uparrow$ &
\textbf{Sent-BERT}~$\uparrow$ &
\textbf{EAD}~$\uparrow$ &
\textbf{Quality}~$\uparrow$ \\
\midrule
\multicolumn{5}{l}{\textbf{LLaDA-8B-Instruct}}\\
\midrule
$\sigma{=}0.1,\ t\in[0.9,0.3]$ & 66.90 & 34.27 & 59.97 & 7.08 \\
$\sigma{=}0.3,\ t\in[0.9,0.3]$ & \textbf{68.64} & \textbf{37.28} & \textbf{60.25} & 6.96 \\
\rowcolor{black!6}
$\sigma{=}0.2,\ t\in[0.9,0.3]$ & 67.80 & 35.51 & 59.58 & 7.05 \\
$\sigma{=}0.2,\ t\in[0.9,0.5]$ & 67.55 & 35.40 & 59.44 & \textbf{7.09} \\
$\sigma{=}0.2,\ t\in[0.5,0.1]$ & 67.10 & 34.22 & 59.90 & 6.99 \\
\midrule
\multicolumn{5}{l}{\textbf{TraDo-8B-Instruct}}\\
\midrule
$\sigma{=}0.1,\ t\in[0.9,0.5]$ & 85.87 & 54.14 & 70.00 & 5.52 \\
$\sigma{=}0.3,\ t\in[0.9,0.5]$ & 85.64 & 54.04 & 70.15 & 5.50 \\
\rowcolor{black!6}
$\sigma{=}0.2,\ t\in[0.9,0.5]$ & 85.83 & \textbf{54.54} & 69.49 & \textbf{5.54} \\
$\sigma{=}0.2,\ t\in[0.9,0.1]$ & \textbf{85.98} & 53.84 & \textbf{70.38} & 5.49 \\
$\sigma{=}0.2,\ t\in[0.5,0.1]$ & 85.71 & 53.62 & 69.59 & 5.46 \\
\bottomrule
\end{tabular}%
}
\end{subtable}
\hfill
\begin{subtable}[t]{0.49\textwidth}
\centering
\caption{Sensitivity to the mixing coefficient $\psi$.}
\label{tab:psi_ablation_llada_singlecol}

\setlength{\tabcolsep}{3pt}
\renewcommand{\arraystretch}{1.24} 
\footnotesize
\resizebox{\linewidth}{!}{%
\begin{tabular}{lccccc}
  \toprule
  \textbf{$\psi$} &
  \textbf{IntraDistinct}~$\uparrow$ &
  \textbf{Div-BLEU}~$\uparrow$ &
  \textbf{Sent-BERT}~$\uparrow$ &
  \textbf{EAD}~$\uparrow$ &
  \textbf{Quality}~$\uparrow$ \\
  \midrule
  \multicolumn{6}{l}{\textbf{LLaDA-8B-Instruct}}\\
  \midrule
  $\psi{=}1.0$ & \textbf{71.94} & 75.43 & 35.90 & \textbf{72.12} & 7.18 \\
  \rowcolor{black!6}
  $\psi{=}0.9$ & 71.73 & 75.62 & 35.44 & 71.82 & \textbf{7.30} \\
  $\psi{=}0.5$ & 71.43 & 75.59 & 36.04 & 71.43 & 7.28 \\
  $\psi{=}0.2$ & 71.54 & \textbf{75.66} & \textbf{35.93} & 72.11 & 7.23 \\
  \midrule
  \multicolumn{6}{l}{\textbf{TraDo-8B-Instruct}}\\
  \midrule
  $\psi{=}1.0$ & \textbf{69.74} & \textbf{85.81} & 54.91 & \textbf{70.19} & 5.45 \\
  \rowcolor{black!6}
  $\psi{=}0.9$ & 69.63 & 85.72 & \textbf{55.15} & 70.06 & \textbf{5.50} \\
  $\psi{=}0.5$ & 69.63 & 85.53 & 54.96 & 69.90 & 5.49 \\
  $\psi{=}0.2$ & 69.55 & 85.63 & 54.70 & 69.96 & 5.48 \\
  \bottomrule
\end{tabular}%
}
\end{subtable}

\end{table*}

\noindent\textbf{Noise scale $\sigma$.}
As shown in Table~\ref{tab:robust_combined_singlecol}, increasing $\sigma$ generally strengthens diversity but introduces a mild quality trade-off. This trend is more pronounced on LLaDA, where larger perturbations improve semantic and embedding-level diversity, while TraDo is less tolerant to strong noise and mainly gains lexical diversity. This suggests that the optimal noise scale depends on backbone-specific denoising dynamics, but moderate perturbations provide a stable diversity--quality balance.

\noindent\textbf{Noise injection time windows $t$.}
Table~\ref{tab:robust_combined_singlecol} shows that injection timing has a stronger effect on semantic diversity. Early-stage perturbations generally yield higher Sent-BERT and EAD scores, while later injection leads to weaker semantic gains and mainly affects lexical realization. This supports our temporal-structure finding that semantic branching is most effective during early denoising, whereas late-stage perturbations can interfere with refinement.



\noindent\textbf{Mixing coefficient $\psi$.}
As shown in Table~\ref{tab:psi_ablation_llada_singlecol}, $\psi$ controls the quality--diversity trade-off by bounding the effective perturbation radius. Fully adopting the perturbed embedding ($\psi=1.0$) yields slightly higher diversity but lower quality, while moderate mixing improves quality with only minor diversity loss. This confirms that pulling the perturbed conditioning back toward the original prompt stabilizes generation without eliminating exploration.

\section{Conclusion}

We propose Time-Annealed Perturbation Sampling (TAPS), a training-free inference strategy for Diffusion-LMs that improves diversity by perturbing the conditioning trajectory in a time-aware and manifold-constrained manner. Our analysis reveals a temporal semantic structure in Diffusion-LMs: early denoising stages are more influential for semantic exploration, while later stages primarily refine local realization. Guided by this structure, TAPS encourages semantic branching early and anneals perturbations away before refinement, improving diversity while preserving prompt alignment, generation quality, and reasoning ability. Experiments across open-ended generation, instruction-following, and verifiable reasoning benchmarks show that TAPS consistently enhances semantic and lexical diversity, supports exploration of alternative reasoning paths, and generalizes across multiple diffusion-based backbones with negligible overhead.

\begin{ack}
This research/project is supported by the National Research Foundation, Singapore under its National Large Language Models Funding Initiative (AISG Award No: AISG-NMLP-2024-003). Any opinions, findings and conclusions or recommendations expressed in this material are those of the author(s) and do not reflect the views of National Research Foundation, Singapore.
\end{ack}

{\small
\bibliographystyle{plainnat}
\bibliography{ref}

\begin{thebibliography}{62}
\providecommand{\natexlab}[1]{#1}
\providecommand{\url}[1]{\texttt{#1}}
\expandafter\ifx\csname urlstyle\endcsname\relax
  \providecommand{\doi}[1]{doi: #1}\else
  \providecommand{\doi}{doi: \begingroup \urlstyle{rm}\Url}\fi

\bibitem[Ackley et~al.(1985)Ackley, Hinton, and Sejnowski]{ackley1985learning}
David~H Ackley, Geoffrey~E Hinton, and Terrence~J Sejnowski.
\newblock A learning algorithm for boltzmann machines.
\newblock \emph{Cognitive science}, 9\penalty0 (1):\penalty0 147--169, 1985.

\bibitem[Basu et~al.(2020)Basu, Ramachandran, Keskar, and Varshney]{basu2020mirostat}
Sourya Basu, Govardana~Sachitanandam Ramachandran, Nitish~Shirish Keskar, and Lav~R Varshney.
\newblock Mirostat: A neural text decoding algorithm that directly controls perplexity.
\newblock \emph{arXiv preprint arXiv:2007.14966}, 2020.

\bibitem[Chang et~al.(2023)Chang, Reitter, Aksitov, and Sung]{chang2023kl}
Chung-Ching Chang, David Reitter, Renat Aksitov, and Yun-Hsuan Sung.
\newblock Kl-divergence guided temperature sampling.
\newblock \emph{arXiv preprint arXiv:2306.01286}, 2023.

\bibitem[Chen et~al.(2025)Chen, Chen, Wang, and Yang]{chen2025seed}
Minghan Chen, Guikun Chen, Wenguan Wang, and Yi~Yang.
\newblock Seed-grpo: Semantic entropy enhanced grpo for uncertainty-aware policy optimization.
\newblock \emph{arXiv preprint arXiv:2505.12346}, 2025.

\bibitem[Cheng et~al.(2025{\natexlab{a}})Cheng, Huang, Zhu, Dai, Zhao, Zhang, and Wei]{cheng2025reasoning}
Daixuan Cheng, Shaohan Huang, Xuekai Zhu, Bo~Dai, Wayne~Xin Zhao, Zhenliang Zhang, and Furu Wei.
\newblock Reasoning with exploration: An entropy perspective.
\newblock \emph{arXiv preprint arXiv:2506.14758}, 2025{\natexlab{a}}.

\bibitem[Cheng et~al.(2025{\natexlab{b}})Cheng, Bian, Liu, Zhang, Yao, Tian, Wang, Guo, Chen, Qi, et~al.]{cheng2025sdar}
Shuang Cheng, Yihan Bian, Dawei Liu, Linfeng Zhang, Qian Yao, Zhongbo Tian, Wenhai Wang, Qipeng Guo, Kai Chen, Biqing Qi, et~al.
\newblock Sdar: A synergistic diffusion-autoregression paradigm for scalable sequence generation.
\newblock \emph{arXiv preprint arXiv:2510.06303}, 2025{\natexlab{b}}.

\bibitem[Chung et~al.(2025)Chung, Padmakumar, Roemmele, Sun, and Kreminski]{chung2025modifying}
John Joon~Young Chung, Vishakh Padmakumar, Melissa Roemmele, Yuqian Sun, and Max Kreminski.
\newblock Modifying large language model post-training for diverse creative writing.
\newblock \emph{arXiv preprint arXiv:2503.17126}, 2025.

\bibitem[Cobbe et~al.(2021)Cobbe, Kosaraju, Bavarian, Chen, Jun, Kaiser, Plappert, Tworek, Hilton, Nakano, Hesse, and Schulman]{cobbe2021gsm8k}
Karl Cobbe, Vineet Kosaraju, Mohammad Bavarian, Mark Chen, Heewoo Jun, Lukasz Kaiser, Matthias Plappert, Jerry Tworek, Jacob Hilton, Reiichiro Nakano, Christopher Hesse, and John Schulman.
\newblock Training verifiers to solve math word problems.
\newblock \emph{arXiv preprint arXiv:2110.14168}, 2021.

\bibitem[Corso et~al.(2023)Corso, Xu, de~Bortoli, Barzilay, and Jaakkola]{corso2023particleguidancenoniiddiverse}
Gabriele Corso, Yilun Xu, Valentin de~Bortoli, Regina Barzilay, and Tommi Jaakkola.
\newblock Particle guidance: non-i.i.d. diverse sampling with diffusion models, 2023.
\newblock URL \url{https://arxiv.org/abs/2310.13102}.

\bibitem[Du et~al.(2024)Du, Winnicki, Dalal, Mannor, and Srikant]{du2024exploration}
Yihan Du, Anna Winnicki, Gal Dalal, Shie Mannor, and R~Srikant.
\newblock Exploration-driven policy optimization in rlhf: Theoretical insights on efficient data utilization.
\newblock \emph{arXiv preprint arXiv:2402.10342}, 2024.

\bibitem[Fan et~al.(2018)Fan, Lewis, and Dauphin]{fan2018hierarchical}
Angela Fan, Mike Lewis, and Yann Dauphin.
\newblock Hierarchical neural story generation.
\newblock \emph{arXiv preprint arXiv:1805.04833}, 2018.

\bibitem[Freitag and Al-Onaizan(2017)]{freitag2017beam}
Markus Freitag and Yaser Al-Onaizan.
\newblock Beam search strategies for neural machine translation.
\newblock \emph{arXiv preprint arXiv:1702.01806}, 2017.

\bibitem[Gruver et~al.(2023)Gruver, Stanton, Frey, Rudner, Hotzel, Lafrance-Vanasse, Rajpal, Cho, and Wilson]{gruver2023protein}
Nate Gruver, Samuel Stanton, Nathan Frey, Tim~GJ Rudner, Isidro Hotzel, Julien Lafrance-Vanasse, Arvind Rajpal, Kyunghyun Cho, and Andrew~G Wilson.
\newblock Protein design with guided discrete diffusion.
\newblock \emph{Advances in neural information processing systems}, 36:\penalty0 12489--12517, 2023.

\bibitem[He et~al.(2024)He, Luo, Bai, Hu, Thai, Shen, Hu, Han, Huang, Zhang, et~al.]{he2024olympiadbench}
Chaoqun He, Renjie Luo, Yuzhuo Bai, Shengding Hu, Zhen Thai, Junhao Shen, Jinyi Hu, Xu~Han, Yujie Huang, Yuxiang Zhang, et~al.
\newblock Olympiadbench: A challenging benchmark for promoting agi with olympiad-level bilingual multimodal scientific problems.
\newblock In \emph{Proceedings of the 62nd Annual Meeting of the Association for Computational Linguistics (Volume 1: Long Papers)}, pages 3828--3850, 2024.

\bibitem[Hendrycks et~al.(2021)Hendrycks, Burns, Kadavath, Arora, Basart, Tang, Song, and Steinhardt]{hendrycks2021measuring}
Dan Hendrycks, Collin Burns, Saurav Kadavath, Akul Arora, Steven Basart, Eric Tang, Dawn Song, and Jacob Steinhardt.
\newblock Measuring mathematical problem solving with the math dataset.
\newblock \emph{arXiv preprint arXiv:2103.03874}, 2021.

\bibitem[Hewitt et~al.(2022)Hewitt, Manning, and Liang]{hewitt2022truncation}
John Hewitt, Christopher~D Manning, and Percy Liang.
\newblock Truncation sampling as language model desmoothing.
\newblock \emph{arXiv preprint arXiv:2210.15191}, 2022.

\bibitem[Holtzman et~al.(2019)Holtzman, Buys, Du, Forbes, and Choi]{holtzman2019curious}
Ari Holtzman, Jan Buys, Li~Du, Maxwell Forbes, and Yejin Choi.
\newblock The curious case of neural text degeneration.
\newblock \emph{arXiv preprint arXiv:1904.09751}, 2019.

\bibitem[Hurst et~al.(2024)Hurst, Lerer, Goucher, Perelman, Ramesh, Clark, Ostrow, Welihinda, Hayes, Radford, et~al.]{hurst2024gpt}
Aaron Hurst, Adam Lerer, Adam~P Goucher, Adam Perelman, Aditya Ramesh, Aidan Clark, AJ~Ostrow, Akila Welihinda, Alan Hayes, Alec Radford, et~al.
\newblock Gpt-4o system card.
\newblock \emph{arXiv preprint arXiv:2410.21276}, 2024.

\bibitem[Ismayilzada et~al.(2025)Ismayilzada, Laverghetta, Luchini, Patel, Bosselut, Van Der~Plas, and Beaty]{ismayilzada2025creative}
Mete Ismayilzada, Antonio Laverghetta, Simone~A Luchini, RN~Patel, Antoine Bosselut, Lonneke Van Der~Plas, and Roger~E Beaty.
\newblock Creative preference optimization.
\newblock \emph{Findings of the Association for Computational Linguistics: EMNLP 2025}, pages 9580--9609, 2025.

\bibitem[Kirk et~al.(2023)Kirk, Mediratta, Nalmpantis, Luketina, Hambro, Grefenstette, and Raileanu]{kirk2023understanding}
Robert Kirk, Ishita Mediratta, Christoforos Nalmpantis, Jelena Luketina, Eric Hambro, Edward Grefenstette, and Roberta Raileanu.
\newblock Understanding the effects of rlhf on llm generalisation and diversity.
\newblock \emph{arXiv preprint arXiv:2310.06452}, 2023.

\bibitem[Lanchantin et~al.(2025)Lanchantin, Chen, Dhuliawala, Yu, Weston, Sukhbaatar, and Kulikov]{lanchantin2025diverse}
Jack Lanchantin, Angelica Chen, Shehzaad Dhuliawala, Ping Yu, Jason Weston, Sainbayar Sukhbaatar, and Ilia Kulikov.
\newblock Diverse preference optimization.
\newblock \emph{arXiv preprint arXiv:2501.18101}, 2025.

\bibitem[Li et~al.(2020{\natexlab{a}})Li, Zhou, He, Wang, Yang, and Li]{li2020sentence}
Bohan Li, Hao Zhou, Junxian He, Mingxuan Wang, Yiming Yang, and Lei Li.
\newblock On the sentence embeddings from pre-trained language models.
\newblock In \emph{Proceedings of the 2020 conference on empirical methods in natural language processing (EMNLP)}, pages 9119--9130, 2020{\natexlab{a}}.

\bibitem[Li et~al.(2016)Li, Galley, Brockett, Gao, and Dolan]{li2016diversity}
Jiwei Li, Michel Galley, Chris Brockett, Jianfeng Gao, and William~B Dolan.
\newblock A diversity-promoting objective function for neural conversation models.
\newblock In \emph{Proceedings of the 2016 conference of the North American chapter of the association for computational linguistics: human language technologies}, pages 110--119, 2016.

\bibitem[Li et~al.(2025{\natexlab{a}})Li, Zhang, Yu, Saha, Khashabi, Weston, Lanchantin, and Wang]{li2025jointly}
Tianjian Li, Yiming Zhang, Ping Yu, Swarnadeep Saha, Daniel Khashabi, Jason Weston, Jack Lanchantin, and Tianlu Wang.
\newblock Jointly reinforcing diversity and quality in language model generations.
\newblock \emph{arXiv preprint arXiv:2509.02534}, 2025{\natexlab{a}}.

\bibitem[Li et~al.(2024{\natexlab{a}})Li, Chiang, Frick, Dunlap, Wu, Zhu, Gonzalez, and Stoica]{li2024crowdsourced}
Tianle Li, Wei-Lin Chiang, Evan Frick, Lisa Dunlap, Tianhao Wu, Banghua Zhu, Joseph~E Gonzalez, and Ion Stoica.
\newblock From crowdsourced data to high-quality benchmarks: Arena-hard and benchbuilder pipeline.
\newblock \emph{arXiv preprint arXiv:2406.11939}, 2024{\natexlab{a}}.

\bibitem[Li et~al.(2025{\natexlab{b}})Li, Chen, Guo, and Shen]{li2025survey}
Tianyi Li, Mingda Chen, Bowei Guo, and Zhiqiang Shen.
\newblock A survey on diffusion language models.
\newblock \emph{arXiv preprint arXiv:2508.10875}, 2025{\natexlab{b}}.

\bibitem[Li et~al.(2022)Li, Thickstun, Gulrajani, Liang, and Hashimoto]{li2022diffusion}
Xiang Li, John Thickstun, Ishaan Gulrajani, Percy~S Liang, and Tatsunori~B Hashimoto.
\newblock Diffusion-lm improves controllable text generation.
\newblock \emph{Advances in neural information processing systems}, 35:\penalty0 4328--4343, 2022.

\bibitem[Li et~al.(2023)Li, Holtzman, Fried, Liang, Eisner, Hashimoto, Zettlemoyer, and Lewis]{li2023contrastive}
Xiang~Lisa Li, Ari Holtzman, Daniel Fried, Percy Liang, Jason Eisner, Tatsunori~B Hashimoto, Luke Zettlemoyer, and Mike Lewis.
\newblock Contrastive decoding: Open-ended text generation as optimization.
\newblock In \emph{Proceedings of the 61st annual meeting of the association for computational linguistics (volume 1: Long papers)}, pages 12286--12312, 2023.

\bibitem[Li et~al.(2024{\natexlab{b}})Li, Chen, Xu, Qin, Xiao, Luo, and Sun]{li2024preserving}
Ziniu Li, Congliang Chen, Tian Xu, Zeyu Qin, Jiancong Xiao, Zhi-Quan Luo, and Ruoyu Sun.
\newblock Preserving diversity in supervised fine-tuning of large language models.
\newblock \emph{arXiv preprint arXiv:2408.16673}, 2024{\natexlab{b}}.

\bibitem[Li et~al.(2020{\natexlab{b}})Li, Wang, Chen, Utiyama, Sumita, Zhang, and Zhao]{li2020data}
Zuchao Li, Rui Wang, Kehai Chen, Masso Utiyama, Eiichiro Sumita, Zhuosheng Zhang, and Hai Zhao.
\newblock Data-dependent gaussian prior objective for language generation.
\newblock In \emph{International Conference on Learning Representations}, 2020{\natexlab{b}}.

\bibitem[Liu et~al.(2024)Liu, Zeng, Liu, Yan, He, Wang, Yan, Liu, and Zhou]{liu2024skywork}
Chris~Yuhao Liu, Liang Zeng, Jiacai Liu, Rui Yan, Jujie He, Chaojie Wang, Shuicheng Yan, Yang Liu, and Yahui Zhou.
\newblock Skywork-reward: Bag of tricks for reward modeling in llms.
\newblock \emph{arXiv preprint arXiv:2410.18451}, 2024.

\bibitem[Liu et~al.(2023)Liu, Xia, Wang, and Zhang]{liu2023your}
Jiawei Liu, Chunqiu~Steven Xia, Yuyao Wang, and Lingming Zhang.
\newblock Is your code generated by chatgpt really correct? rigorous evaluation of large language models for code generation.
\newblock \emph{Advances in neural information processing systems}, 36:\penalty0 21558--21572, 2023.

\bibitem[Liu et~al.(2022)Liu, Sabour, Zheng, Ke, Zhu, and Huang]{liu2022rethinking}
Siyang Liu, Sahand Sabour, Yinhe Zheng, Pei Ke, Xiaoyan Zhu, and Minlie Huang.
\newblock Rethinking and refining the distinct metric.
\newblock \emph{arXiv preprint arXiv:2202.13587}, 2022.

\bibitem[Mai and Carson-Berndsen(2024)]{mai2024improving}
Long Mai and Julie Carson-Berndsen.
\newblock Improving linguistic diversity of large language models with possibility exploration fine-tuning.
\newblock \emph{arXiv preprint arXiv:2412.03343}, 2024.

\bibitem[Nagarajan et~al.(2025)Nagarajan, Wu, Ding, and Raghunathan]{nagarajan2025roll}
Vaishnavh Nagarajan, Chen~Henry Wu, Charles Ding, and Aditi Raghunathan.
\newblock Roll the dice \& look before you leap: Going beyond the creative limits of next-token prediction.
\newblock \emph{arXiv preprint arXiv:2504.15266}, 2025.

\bibitem[Nguyen et~al.(2024)Nguyen, Baker, Neo, Roush, Kirsch, and Shwartz-Ziv]{nguyen2024turning}
Minh~Nhat Nguyen, Andrew Baker, Clement Neo, Allen Roush, Andreas Kirsch, and Ravid Shwartz-Ziv.
\newblock Turning up the heat: Min-p sampling for creative and coherent llm outputs.
\newblock \emph{arXiv preprint arXiv:2407.01082}, 2024.

\bibitem[Nie et~al.(2025)Nie, Zhu, You, Zhang, Ou, Hu, Zhou, Lin, Wen, and Li]{nie2025large}
Shen Nie, Fengqi Zhu, Zebin You, Xiaolu Zhang, Jingyang Ou, Jun Hu, Jun Zhou, Yankai Lin, Ji-Rong Wen, and Chongxuan Li.
\newblock Large language diffusion models.
\newblock \emph{arXiv preprint arXiv:2502.09992}, 2025.

\bibitem[Novikov et~al.(2025)Novikov, V{\~u}, Eisenberger, Dupont, Huang, Wagner, Shirobokov, Kozlovskii, Ruiz, Mehrabian, et~al.]{novikov2025alphaevolve}
Alexander Novikov, Ng{\^a}n V{\~u}, Marvin Eisenberger, Emilien Dupont, Po-Sen Huang, Adam~Zsolt Wagner, Sergey Shirobokov, Borislav Kozlovskii, Francisco~JR Ruiz, Abbas Mehrabian, et~al.
\newblock Alphaevolve: A coding agent for scientific and algorithmic discovery.
\newblock \emph{arXiv preprint arXiv:2506.13131}, 2025.

\bibitem[Pan et~al.(2026)Pan, Wan, Yu, Zheng, Ke, Sun, Wang, Wang, and Tsang]{pan2026savlaspatiallyawareflowmatchingvisionlanguageaction}
Xu~Pan, Zhenglin Wan, Xingrui Yu, Xianwei Zheng, Youkai Ke, Ming Sun, Rui Wang, Ziwei Wang, and Ivor Tsang.
\newblock Sa-vla: Spatially-aware flow-matching for vision-language-action reinforcement learning, 2026.
\newblock URL \url{https://arxiv.org/abs/2602.00743}.

\bibitem[Papineni et~al.(2002)Papineni, Roukos, Ward, and Zhu]{papineni2002bleu}
Kishore Papineni, Salim Roukos, Todd Ward, and Wei-Jing Zhu.
\newblock Bleu: a method for automatic evaluation of machine translation.
\newblock In \emph{Proceedings of the 40th annual meeting of the Association for Computational Linguistics}, pages 311--318, 2002.

\bibitem[Peeperkorn et~al.(2024)Peeperkorn, Kouwenhoven, Brown, and Jordanous]{peeperkorn2024temperature}
Max Peeperkorn, Tom Kouwenhoven, Dan Brown, and Anna Jordanous.
\newblock Is temperature the creativity parameter of large language models?
\newblock \emph{arXiv preprint arXiv:2405.00492}, 2024.

\bibitem[Rein et~al.(2023)Rein, Hou, Stickland, Petty, Pang, Dirani, Michael, and Bowman]{rein2023gpqa}
David Rein, Betty~Li Hou, Asa~Cooper Stickland, Jackson Petty, Richard~Yuanzhe Pang, Julien Dirani, Julian Michael, and Samuel~R Bowman.
\newblock Gpqa: A graduate-level google-proof q\&a benchmark.
\newblock \emph{arXiv preprint arXiv:2311.12022}, 2023.

\bibitem[Ruan et~al.(2025)Ruan, Li, Liu, Chen, Luo, Li, Liu, and Chen]{ruan2025g2}
Zhiwen Ruan, Yixia Li, Yefeng Liu, Yun Chen, Weihua Luo, Peng Li, Yang Liu, and Guanhua Chen.
\newblock G2: Guided generation for enhanced output diversity in llms.
\newblock In \emph{Proceedings of the 2025 Conference on Empirical Methods in Natural Language Processing}, pages 14127--14145, 2025.

\bibitem[Sadat et~al.(2023)Sadat, Buhmann, Bradley, Hilliges, and Weber]{sadat2023cads}
Seyedmorteza Sadat, Jakob Buhmann, Derek Bradley, Otmar Hilliges, and Romann~M Weber.
\newblock Cads: Unleashing the diversity of diffusion models through condition-annealed sampling.
\newblock \emph{arXiv preprint arXiv:2310.17347}, 2023.

\bibitem[Si et~al.(2024)Si, Yang, and Hashimoto]{si2024llmsgeneratenovelresearch}
Chenglei Si, Diyi Yang, and Tatsunori Hashimoto.
\newblock Can llms generate novel research ideas? a large-scale human study with 100+ nlp researchers, 2024.
\newblock URL \url{https://arxiv.org/abs/2409.04109}.

\bibitem[Slocum et~al.(2025)Slocum, Parker-Sartori, and Hadfield-Menell]{slocum2025diverse}
Stewart Slocum, Asher Parker-Sartori, and Dylan Hadfield-Menell.
\newblock Diverse preference learning for capabilities and alignment.
\newblock In \emph{The Thirteenth International Conference on Learning Representations}, 2025.

\bibitem[Stiennon et~al.(2020)Stiennon, Ouyang, Wu, Ziegler, Lowe, Voss, Radford, Amodei, and Christiano]{stiennon2020learning}
Nisan Stiennon, Long Ouyang, Jeffrey Wu, Daniel Ziegler, Ryan Lowe, Chelsea Voss, Alec Radford, Dario Amodei, and Paul~F Christiano.
\newblock Learning to summarize with human feedback.
\newblock \emph{Advances in neural information processing systems}, 33:\penalty0 3008--3021, 2020.

\bibitem[Tae et~al.(2025)Tae, Ivison, Kumar, and Cohan]{tae2025tess}
Jaesung Tae, Hamish Ivison, Sachin Kumar, and Arman Cohan.
\newblock Tess 2: A large-scale generalist diffusion language model.
\newblock \emph{arXiv preprint arXiv:2502.13917}, 2025.

\bibitem[Wan et~al.(2025)Wan, Yu, Bossens, Lyu, Guo, Fan, Ong, and Tsang]{wan2025diversifyingpolicybehaviorsextrinsic}
Zhenglin Wan, Xingrui Yu, David~Mark Bossens, Yueming Lyu, Qing Guo, Flint~Xiaofeng Fan, Yew~Soon Ong, and Ivor Tsang.
\newblock Diversifying policy behaviors with extrinsic behavioral curiosity, 2025.
\newblock URL \url{https://arxiv.org/abs/2410.06151}.

\bibitem[Wang et~al.(2025{\natexlab{a}})Wang, Yu, Gao, Zheng, Liu, Lu, Dang, Chen, Yang, Zhang, et~al.]{wang2025beyond}
Shenzhi Wang, Le~Yu, Chang Gao, Chujie Zheng, Shixuan Liu, Rui Lu, Kai Dang, Xionghui Chen, Jianxin Yang, Zhenru Zhang, et~al.
\newblock Beyond the 80/20 rule: High-entropy minority tokens drive effective reinforcement learning for llm reasoning.
\newblock \emph{arXiv preprint arXiv:2506.01939}, 2025{\natexlab{a}}.

\bibitem[Wang et~al.(2025{\natexlab{b}})Wang, Yang, Li, Tian, Shen, and Wang]{wang2025revolutionizing}
Yinjie Wang, Ling Yang, Bowen Li, Ye~Tian, Ke~Shen, and Mengdi Wang.
\newblock Revolutionizing reinforcement learning framework for diffusion large language models.
\newblock \emph{arXiv preprint arXiv:2509.06949}, 2025{\natexlab{b}}.

\bibitem[Welleck et~al.(2019)Welleck, Kulikov, Roller, Dinan, Cho, and Weston]{welleck2019neural}
Sean Welleck, Ilia Kulikov, Stephen Roller, Emily Dinan, Kyunghyun Cho, and Jason Weston.
\newblock Neural text generation with unlikelihood training.
\newblock \emph{arXiv preprint arXiv:1908.04319}, 2019.

\bibitem[Wu et~al.(2025{\natexlab{a}})Wu, Zhang, Xue, Liu, Diao, Zhu, Luo, Han, and Xie]{wu2025fast}
Chengyue Wu, Hao Zhang, Shuchen Xue, Zhijian Liu, Shizhe Diao, Ligeng Zhu, Ping Luo, Song Han, and Enze Xie.
\newblock Fast-dllm: Training-free acceleration of diffusion llm by enabling kv cache and parallel decoding.
\newblock \emph{arXiv preprint arXiv:2505.22618}, 2025{\natexlab{a}}.

\bibitem[Wu et~al.(2025{\natexlab{b}})Wu, Wan, Yu, Yang, An, and Tsang]{wu2025oscarorthogonalstochasticcontrol}
Jingxuan Wu, Zhenglin Wan, Xingrui Yu, Yuzhe Yang, Bo~An, and Ivor Tsang.
\newblock Oscar: Orthogonal stochastic control for alignment-respecting diversity in flow matching, 2025{\natexlab{b}}.
\newblock URL \url{https://arxiv.org/abs/2510.09060}.

\bibitem[Ye et~al.(2025)Ye, Xie, Zheng, Gao, Wu, Jiang, Li, and Kong]{ye2025dream}
Jiacheng Ye, Zhihui Xie, Lin Zheng, Jiahui Gao, Zirui Wu, Xin Jiang, Zhenguo Li, and Lingpeng Kong.
\newblock Dream 7b: Diffusion large language models.
\newblock \emph{arXiv preprint arXiv:2508.15487}, 2025.

\bibitem[Yu et~al.(2025)Yu, Wan, Bossens, Lyu, Guo, and Tsang]{yu2025imitationdiversebehaviorswasserstein}
Xingrui Yu, Zhenglin Wan, David~Mark Bossens, Yueming Lyu, Qing Guo, and Ivor~W. Tsang.
\newblock Imitation from diverse behaviors: Wasserstein quality diversity imitation learning with single-step archive exploration, 2025.
\newblock URL \url{https://arxiv.org/abs/2411.06965}.

\bibitem[Zhang et~al.(2026)Zhang, Wan, Chen, Yang, Feng, Zhou, Yu, You, Tsang, and An]{zhang2026evolvingdiffusionflowmatching}
Chubin Zhang, Zhenglin Wan, Feng Chen, Fuchao Yang, Lang Feng, Yaxin Zhou, Xingrui Yu, Yang You, Ivor Tsang, and Bo~An.
\newblock Evolving diffusion and flow matching policies for online reinforcement learning, 2026.
\newblock URL \url{https://arxiv.org/abs/2512.02581}.

\bibitem[Zhang et~al.(2024{\natexlab{a}})Zhang, Bao, and Huang]{zhang2024edt}
Shimao Zhang, Yu~Bao, and Shujian Huang.
\newblock Edt: Improving large language models' generation by entropy-based dynamic temperature sampling.
\newblock \emph{arXiv preprint arXiv:2403.14541}, 2024{\natexlab{a}}.

\bibitem[Zhang et~al.(2024{\natexlab{b}})Zhang, Schwarzschild, Carlini, Kolter, and Ippolito]{zhang2024forcing}
Yiming Zhang, Avi Schwarzschild, Nicholas Carlini, Zico Kolter, and Daphne Ippolito.
\newblock Forcing diffuse distributions out of language models.
\newblock \emph{arXiv preprint arXiv:2404.10859}, 2024{\natexlab{b}}.

\bibitem[Zhang et~al.(2025)Zhang, Diddee, Holm, Liu, Liu, Samuel, Wang, and Ippolito]{zhang2025noveltybench}
Yiming Zhang, Harshita Diddee, Susan Holm, Hanchen Liu, Xinyue Liu, Vinay Samuel, Barry Wang, and Daphne Ippolito.
\newblock Noveltybench: Evaluating language models for humanlike diversity.
\newblock \emph{arXiv preprint arXiv:2504.05228}, 2025.

\bibitem[Zhu et~al.(2018)Zhu, Lu, Zheng, Guo, Zhang, Wang, and Yu]{zhu2018texygen}
Yaoming Zhu, Sidi Lu, Lei Zheng, Jiaxian Guo, Weinan Zhang, Jun Wang, and Yong Yu.
\newblock Texygen: A benchmarking platform for text generation models.
\newblock In \emph{The 41st international ACM SIGIR conference on research \& development in information retrieval}, pages 1097--1100, 2018.

\bibitem[Zhu et~al.(2024)Zhu, Li, Li, Zhao, Jin, and Mei]{zhu2024hot}
Yuqi Zhu, Jia Li, Ge~Li, YunFei Zhao, Zhi Jin, and Hong Mei.
\newblock Hot or cold? adaptive temperature sampling for code generation with large language models.
\newblock In \emph{Proceedings of the AAAI Conference on Artificial Intelligence}, volume~38, pages 437--445, 2024.

\end{thebibliography}
}


\newpage
\appendix

\section{Implementation Details}
\label{sec:imple_detail}

\subsection{Hyperparameter Settings}
\label{sec:hyper_set}


We use backbone-specific hyperparameter configurations for LLaDA-8B-Instruct, TraDo-8B-Instruct, SDAR-8B-Chat, and Dream-v0-Instruct-7B, following their respective default decoding setups. Differences in denoising steps, block length, remasking strategy, and sampling algorithm reflect architectural variations across non-autoregressive, semi-autoregressive, and block-wise diffusion language models. Unless otherwise specified, these backbone hyperparameters are fixed across all experiments. Detailed configurations are provided in Table~\ref{tab:backbone_hparams_all}. In addition, TAPS uses a small set of backbone-specific settings while keeping its design unchanged across tasks and temperature settings, supporting the robustness and generality of the proposed approach.

\begin{table}[!h]
\centering
\small
\setlength{\tabcolsep}{5pt}
\renewcommand{\arraystretch}{1.12}
\caption{Backbone-specific generation and TAPS hyperparameter settings. ``--'' indicates that the hyperparameter is not applicable to the corresponding backbone.}
\label{tab:backbone_hparams_all}
\resizebox{\linewidth}{!}{%
\begin{tabular}{lcccc}
\toprule
\textbf{Hyperparameter} 
& \textbf{LLaDA-8B-Instruct} 
& \textbf{TraDo-8B-Instruct} 
& \textbf{SDAR-8B-Chat} 
& \textbf{Dream-v0-Instruct-7B} \\
\midrule
\multicolumn{5}{l}{\textbf{Backbone Settings}} \\
\midrule
Mask ID & 126336 & 151669 & 151669 & 151666 \\
Generation Length & 256 & 200 & 256 & 256 \\
Block Length & 128 & 4 & 4 & -- \\
Denoising Steps & 256 & 4 & 4 & 256 \\
CFG Scale & 0.0 & -- & -- & -- \\
Remasking Strategy & low\_confidence & low\_confidence\_dynamic & low\_confidence\_dynamic & -- \\
Confidence Threshold & -- & 0.9 & 0.85 & -- \\
Sampling Algorithm & -- & -- & -- & origin \\
Algorithm Temperature & -- & -- & -- & 0.0 \\
\midrule
\multicolumn{5}{l}{\textbf{TAPS Settings}} \\
\midrule
Noise Scale $\sigma$ & 0.20 & 0.20 & 0.10 & 0.20 \\
Noise Injection Window $t$ & [0.90, 0.30] & [0.90, 0.50] & [0.90, 0.30] & [0.90, 0.30] \\
Noise Rescale Coefficient $\psi$ & 0.90 & 0.90 & 0.90 & 0.90 \\
Annealing Strategy & cosine & cosine & linear & cosine \\
\bottomrule
\end{tabular}%
}
\end{table}

\subsection{Backbone-Specific Generation Settings}

Although our method is model-agnostic in principle, we adopt different generation configurations for LLaDA and TraDo due to architectural and practical constraints.

TraDo is trained on SDAR weights and follows a block-wise masked generation paradigm \citep{cheng2025sdar}, which adapts an autoregressive Transformer language model. In practice, we find that the block length in TraDo must be set to a small value, specifically, 4 tokens in the official GitHub. Increasing the block length leads to severe generation instability, including garbled outputs or empty generations. Consequently, generating a sequence of 256 tokens requires 64 blocks. Under this setting, diffusion steps within each block are necessarily limited, and semantic construction and surface-level refinement are tightly coupled within a very short horizon. To accommodate this constraint, we inject context noise across blocks with a gradually decaying schedule, allowing early blocks to introduce diversity while later blocks stabilize the output.

In contrast, LLaDA supports substantially larger block lengths. In our experiments, we set the block length to 128 tokens and use a larger number of denoising steps within each block. This configuration more closely aligns with the intended behavior of diffusion language models, where early denoising steps primarily determine high-level semantic structure, and later steps focus on lexical choice and local fluency. As a result, context perturbations injected early in the diffusion process can more effectively induce semantic diversity without significantly harming generation quality. Overall, the greater flexibility in block length and denoising depth allows our method to better exploit the temporal structure of diffusion in LLaDA, which partly explains its stronger empirical performance compared to TraDo.

\subsection{Baseline Implementation}
\label{sec:baseline_implementation}
We implement all baseline decoding methods under a unified evaluation framework. For sampling-based methods, we tune their key hyperparameters over commonly used ranges and report the best-performing configuration in the main results. Specifically, for Top-$k$, we search over $\{30, 40, 50\}$; for Top-$p$, we consider $\{0.70, 0.90, 0.95\}$; and for Min-$p$, we use $\{0.10, 0.20, 0.30\}$. 

For Diverse Prompt, we follow its original implementation and use the recommended configuration, including a cosine-annealed conditional noise schedule with noise injected from early to late denoising steps ($t\in[0.05, 0.95]$), a prompt masking ratio of $0.05$, and an embedding noise standard deviation of $0.20$.

For EDT, we adopt the official implementation, where a dynamic temperature is computed based on token-level entropy. We use the default settings, including base temperature $t_0=0.7$, scaling factor $\theta=20.0$, and $n_{\text{const}}=0.8$, with numerical stability parameter $\epsilon=10^{-6}$. Following the original setup, the base temperature is fixed to $1.0$ when EDT is applied, as temperature scaling is handled internally.

All baselines are applied to the same pretrained backbone models without additional fine-tuning. Unless otherwise specified, all methods use identical decoding configurations, ensuring that performance differences arise solely from the decoding strategy rather than implementation or configuration discrepancies.

\subsection{Evaluation Details}
\label{sec:eval_detail}

We evaluate diversity using lexical- and semantic-level metrics computed from multiple generations per prompt. For each prompt, we first apply light text cleaning by removing template/special tokens (e.g., patterns like \texttt{<|...|>} and short XML-like tags) and normalizing whitespace. We then discard generations shorter than a minimum character threshold and keep at most the top $N$ longest remaining samples per prompt; prompts with fewer than two valid samples are excluded from evaluation. Tokenization for lexical metrics follows a simple regex-based tokenizer that splits text into word tokens and punctuation marks.

IntraDistinct is computed as the average distinct-$n$ ratio over samples, with $n\in\{1,2,3\}$:
\[
\mathrm{Distinct}_n(x)=\frac{|\mathrm{uniq}(\mathrm{ngram}_n(x))|}{|\mathrm{ngram}_n(x)|},\quad
\mathrm{IntraDistinct}=\frac{1}{3}\sum_{n=1}^{3}\mathbb{E}_{x}\big[\mathrm{Distinct}_n(x)\big],
\]
where $\mathrm{ngram}_n(x)$ denotes sets of $n$-grams in generation $x$. We also report EAD following an occupancy-style normalization: for a set of generations $\mathcal{X}$ of a prompt, let $C_n$ be the total number of extracted $n$-grams across $\mathcal{X}$ and $N_n$ be the number of unique $n$-grams; with global vocabulary size $V_n$ computed from all evaluated generations, we define
\[
\mathrm{EAD}_n(\mathcal{X})=\frac{N_n}{V_n\left(1-\left(\frac{V_n-1}{V_n}\right)^{C_n}\right)}.
\]
We average $\mathrm{EAD}_n$ over $n=1,\ldots,5$ to obtain a single EAD score scaled by $100$ for presentation. 

For semantic diversity, we use Sentence-BERT embeddings from the \texttt{all-MiniLM-L6-v2} checkpoint. We encode all generations for the same prompt, L2-normalize the embeddings, and compute the mean pairwise cosine distance over all upper-triangular pairs; the resulting Sent-BERT diversity score is scaled by $100$. All metrics are first computed per prompt and then averaged across prompts to obtain dataset-level results.

\paragraph{Diversity of correct reasoning paths.}
\label{sec:path_metric_details}
For Table~\ref{tab:correct_path_diversity}, we use a shared subset of 100 GSM8K questions and 60 HumanEval+ problems on which every method produces at least three correct outputs out of five. We discard incorrect outputs and measure diversity among the remaining correct solutions separately for each method and problem.

For HumanEval+, \emph{Norm-AST Diff.} compares the structures of generated programs after abstract-syntax-tree normalization, reducing differences that do not change program structure. \emph{Alg. Skeleton} compares the algorithmic structures of the correct programs at a coarser level. \emph{Identical} measures repeated program structures; lower values indicate fewer duplicate correct solutions.

For GSM8K, \emph{Distinct-path} measures variation in the reasoning paths leading to the correct answer. \emph{Op. Edit} measures differences in the mathematical-operation sequences used by those paths. \emph{Identical} measures repeated reasoning paths; lower values indicate fewer duplicate correct solutions. These metrics complement answer-level accuracy by assessing variation among solutions that reach a correct result.

\section{Algorithm}
\label{sec:algo}

This section presents the complete TAPS Algorithm~\ref{alg:taps}, including embedding-level conditioning perturbation, noise annealing, and the associated quality preservation mechanisms.

\begin{algorithm}[h]
\caption{TAPS (Time-Annealed Perturbation Sampling)}
\label{alg:taps}
\KwRequire{Diffusion LM $f_\theta$; prompt token ids $\mathbf{p}$; mask token id $m$; total steps $S$; generation length $L$; block length $B$; noise window $[t_{\text{start}}, t_{\text{end}}]$; annealing schedule $\sigma(t)$ with maximum scale $\sigma_{\max}$; mixing coefficient $\psi \in [0,1]$.}
\KwOutput{Completed sequence $\mathbf{x}\in \mathbb{N}^{|\mathbf{p}|+L}$.}

$\mathbf{x} \gets [\mathbf{p}, \underbrace{m,\dots,m}_{L}]$\;
$\mathbf{E} \gets \mathrm{Embed}(\mathbf{p}) \in \mathbb{R}^{T\times d}$\;
$N \gets L/B$\;
$S_b \gets \lfloor S / N \rfloor$\;
$S' \gets N \cdot S_b$\;

\For{$b \gets 0$ \KwTo $N-1$}{
  Let $\mathcal{I}_b$ be the token indices of the current block\;
  Precompute transfer counts $\{k_i\}_{i=1}^{S_b}$ from the block mask pattern\;

  \For{$i \gets 1$ \KwTo $S_b$}{
    $g \gets b \cdot S_b + i$\;
    $t \gets g / S'$\;

    \eIf{$t \in [t_{\text{start}}, t_{\text{end}}]$}{
      $\tilde{\mathbf{E}} \gets \mathbf{E} + \sigma(t)\sigma_{\max}\boldsymbol{\epsilon},\ 
      \boldsymbol{\epsilon} \sim \mathcal{N}(\mathbf{0}, \mathbf{I})$\tcp*[r]{Sample nearby conditioning trajectory}

      Distribution-aware rescale $\tilde{\mathbf{E}}$ to match statistics of $\mathbf{E}$\;

      $\mathbf{E}' \gets \psi\,\tilde{\mathbf{E}} + (1-\psi)\,\mathbf{E}$\tcp*[r]{Bound perturbation radius}

      \For{$j \gets 1$ \KwTo $T$}{
        $\mathbf{E}'_j \gets 
        \dfrac{\mathbf{E}'_j}{\|\mathbf{E}'_j\|_2 + \varepsilon}
        \cdot \|\mathbf{E}_j\|_2$\tcp*[r]{Preserve token-wise geometry}
      }
    }{
      $\mathbf{E}' \gets \mathbf{E}$\tcp*[r]{Recover original conditioning}
    }

    $\mathbf{z} \gets f_\theta(\mathbf{x};\mathbf{E}')$\;
    $\mathbf{x}_0 \gets \arg\max(\mathbf{z} + \mathrm{Gumbel}(\tau))$\;
    Compute confidence scores $\mathbf{c}$ from $\mathbf{z}$\;
    Select $k_i$ masked positions in $\mathcal{I}_b$ with lowest confidence\;
    Update selected positions in $\mathbf{x}$ using $\mathbf{x}_0$\;
  }
}
\Return $\mathbf{x}$\;
\end{algorithm}

\section{Theoretical Analysis of Bounded Trajectory Deviation}
\label{sec:theory_taps}

We provide a simple theoretical analysis showing that TAPS induces a bounded deviation from the standard denoising trajectory. 
The goal is not to prove exact equivalence between TAPS and standard inference, but to show that the deviation introduced by conditioning perturbations is controlled by a finite perturbation budget. 
This analysis formalizes the intuition that early perturbations allow TAPS to explore nearby semantic trajectories, while annealing the perturbation to zero limits its effect on the final refinement stage.

\paragraph{Denoising dynamics.}
Let $\mathbf{z}_t$ denote the denoising state at step $t$, and let $\mathbf{E}$ be the original prompt embedding. 
We write one step of standard Diffusion-LM inference as
\[
\mathbf{z}_{t-1}=F_t(\mathbf{z}_t,\mathbf{E}),
\]
where $F_t$ denotes the denoising update at step $t$. 
Under TAPS, the update becomes
\[
\tilde{\mathbf{z}}_{t-1}=F_t(\tilde{\mathbf{z}}_t,\mathbf{E}^{\ast}_t),
\]
where $\mathbf{E}^{\ast}_t$ is the manifold-constrained perturbed conditioning embedding. 
Outside the active perturbation window, TAPS recovers the original conditioning signal, i.e., $\mathbf{E}^{\ast}_t=\mathbf{E}$.

\begin{assumption}[Local Lipschitz denoising dynamics]
\label{assump:lipschitz}
For each denoising step $t$, the update map $F_t$ is locally Lipschitz with respect to both the denoising state and the conditioning embedding. 
That is, there exist constants $L_z,L_E>0$ such that
\[
\|F_t(\mathbf{z},\mathbf{E})-F_t(\mathbf{z}',\mathbf{E}')\|
\leq
L_z\|\mathbf{z}-\mathbf{z}'\| + L_E\|\mathbf{E}-\mathbf{E}'\|.
\]
\end{assumption}

\begin{assumption}[Bounded projected conditioning perturbation]
\label{assump:bounded_conditioning}
After distribution-aware rescaling, mixing with the original conditioning, and norm-preserving projection, the effective conditioning perturbation used by TAPS satisfies
\[
\|\mathbf{E}^{\ast}_t-\mathbf{E}\|
\leq
C_E \psi \sigma(t)
\]
for some constant $C_E>0$, where $\psi$ is the mixing coefficient and $\sigma(t)$ is the annealed perturbation scale. 
For steps outside the active perturbation window, $\sigma(t)=0$ and therefore $\mathbf{E}^{\ast}_t=\mathbf{E}$.
\end{assumption}

Assumption~\ref{assump:bounded_conditioning} captures the effect of the manifold-constrained perturbation operations used in TAPS. 
The Gaussian noise is not assumed to be deterministically bounded in isolation; rather, the assumption applies to the effective perturbation after rescaling, mixing, projection, and restriction to the local neighborhood explored by TAPS.

\begin{theorem}[Trajectory deviation bound]
\label{thm:trajectory_deviation}
Let $\mathbf{z}_t$ and $\tilde{\mathbf{z}}_t$ be the trajectories generated by standard inference and TAPS, respectively, initialized from the same state $\mathbf{z}_T=\tilde{\mathbf{z}}_T$. 
Under Assumptions~\ref{assump:lipschitz} and~\ref{assump:bounded_conditioning}, the final deviation satisfies
\[
\|\tilde{\mathbf{z}}_0-\mathbf{z}_0\|
\leq
L_E C_E \psi
\sum_{t=1}^{T} L_z^{t-1}\sigma(t).
\]
In particular, if $L_z\leq 1$, then
\[
\|\tilde{\mathbf{z}}_0-\mathbf{z}_0\|
\leq
L_E C_E \psi
\sum_{t=1}^{T}\sigma(t).
\]
\end{theorem}

\begin{proof}
Define the trajectory deviation
\[
\Delta_t=\|\tilde{\mathbf{z}}_t-\mathbf{z}_t\|.
\]
Using the update rules and Assumption~\ref{assump:lipschitz}, we have
\[
\begin{aligned}
\Delta_{t-1}
&=
\|F_t(\tilde{\mathbf{z}}_t,\mathbf{E}^{\ast}_t)
-
F_t(\mathbf{z}_t,\mathbf{E})\| \\
&\leq
L_z\|\tilde{\mathbf{z}}_t-\mathbf{z}_t\|
+
L_E\|\mathbf{E}^{\ast}_t-\mathbf{E}\| \\
&=
L_z\Delta_t
+
L_E\|\mathbf{E}^{\ast}_t-\mathbf{E}\|.
\end{aligned}
\]
By Assumption~\ref{assump:bounded_conditioning},
\[
\Delta_{t-1}
\leq
L_z\Delta_t + L_E C_E\psi\sigma(t).
\]
Since the two trajectories are initialized from the same state, $\Delta_T=0$. 
Recursively unrolling the above inequality gives
\[
\Delta_0
\leq
L_E C_E \psi
\sum_{t=1}^{T} L_z^{t-1}\sigma(t).
\]
If $L_z\leq 1$, then $L_z^{t-1}\leq 1$ for all $t$, and therefore
\[
\Delta_0
\leq
L_E C_E \psi
\sum_{t=1}^{T}\sigma(t).
\]
This proves the result.
\end{proof}

Theorem~\ref{thm:trajectory_deviation} shows that the deviation induced by TAPS is controlled by the accumulated perturbation budget
\[
B=\psi\sum_{t=1}^{T}\sigma(t).
\]
Since TAPS uses a monotonically decreasing schedule and sets $\sigma(t)=0$ outside the active perturbation window, the accumulated deviation remains finite and explicitly controllable. 
This supports the design principle of TAPS: early perturbations enable exploration of alternative semantic or reasoning trajectories, while later recovery to the original conditioning signal helps preserve prompt alignment, generation quality, and reasoning fidelity.

\begin{corollary}[Bounded distributional shift]
\label{cor:distribution_shift}
Let $P_{\mathrm{std}}$ and $P_{\mathrm{TAPS}}$ denote the output distributions induced by standard inference and TAPS, respectively. 
Suppose the final prediction map from denoising state to output distribution is locally Lipschitz under a distributional distance $D(\cdot,\cdot)$, i.e.,
\[
D\big(P(\cdot \mid \tilde{\mathbf{z}}_0), P(\cdot \mid \mathbf{z}_0)\big)
\leq
L_P \|\tilde{\mathbf{z}}_0-\mathbf{z}_0\|
\]
for some constant $L_P>0$. 
Then, under the assumptions of Theorem~\ref{thm:trajectory_deviation}, the distributional shift induced by TAPS is bounded by
\[
D(P_{\mathrm{TAPS}}, P_{\mathrm{std}})
\leq
L_P L_E C_E \psi
\sum_{t=1}^{T} L_z^{t-1}\sigma(t).
\]
In particular, if $L_z \leq 1$, then
\[
D(P_{\mathrm{TAPS}}, P_{\mathrm{std}})
\leq
L_P L_E C_E \psi
\sum_{t=1}^{T}\sigma(t).
\]
\end{corollary}

\begin{proof}
By the local Lipschitz assumption on the final prediction map,
\[
\begin{aligned}
D(P_{\mathrm{TAPS}}, P_{\mathrm{std}})
&=
D\big(P(\cdot \mid \tilde{\mathbf{z}}_0), P(\cdot \mid \mathbf{z}_0)\big) \\
&\leq
L_P \|\tilde{\mathbf{z}}_0-\mathbf{z}_0\|.
\end{aligned}
\]
Applying the trajectory deviation bound from Theorem~\ref{thm:trajectory_deviation} gives
\[
D(P_{\mathrm{TAPS}}, P_{\mathrm{std}})
\leq
L_P L_E C_E \psi
\sum_{t=1}^{T} L_z^{t-1}\sigma(t).
\]
If $L_z \leq 1$, this further reduces to
\[
D(P_{\mathrm{TAPS}}, P_{\mathrm{std}})
\leq
L_P L_E C_E \psi
\sum_{t=1}^{T}\sigma(t).
\]
Thus, the distributional shift introduced by TAPS is controlled by the same accumulated perturbation budget that bounds trajectory deviation.
\end{proof}

\section{Additional Experiment Results}
\label{sec:add_exp}

\subsection{Existing Assets and Licenses}
\label{sec:assets_licenses}

We use existing datasets, benchmarks, pretrained models, and evaluation models in this work. 
We cite the original sources of these assets and use them only for research evaluation. 
Table~\ref{tab:assets_licenses} summarizes the assets used in this paper, their roles, and the corresponding licenses or terms of use where available.

\begin{table}[t]
\centering
\caption{Existing assets used in this work.}
\label{tab:assets_licenses}
\small
\begin{tabular}{llll}
\toprule
Asset & Type & Use in this paper & License / terms \\
\midrule
\multicolumn{4}{l}{\textbf{Datasets \& benchmarks}} \\
WritingPrompts & Dataset & Story generation evaluation & MIT License \\
NoveltyBench & Benchmark & Multi-sample diversity evaluation & MIT License \\
Arena-Hard-Auto & Benchmark & Preference-aligned creative writing & Apache-2.0 License \\
GSM8K & Dataset & Mathematical reasoning evaluation & MIT License \\
MATH500 & Benchmark & Mathematical reasoning evaluation & MIT License \\
GPQA-Diamond & Benchmark & Scientific reasoning evaluation & MIT License \\
HumanEval+ & Benchmark & Code generation evaluation & MIT License \\
\midrule
\multicolumn{4}{l}{\textbf{Backbone models}} \\
LLaDA-8B-Instruct & Model & Diffusion-LM backbone & MIT License \\
Dream-Instruct-7B & Model & Diffusion-LM backbone & Apache-2.0 License \\
SDAR-8B-Chat & Model & Diffusion-LM backbone & MIT License \\
TraDo-8B-Instruct & Model & Diffusion-LM backbone & Apache-2.0 License \\
\midrule
\multicolumn{4}{l}{\textbf{Evaluation models}} \\
Skywork-Reward & Model & Automatic quality evaluation & Skywork Community License \\
GPT-4o & API model & Preference / quality evaluation & OpenAI Services Agreement \\
Gemini-2 & API model & Preference evaluation & Gemini API Terms of Service \\
\bottomrule
\end{tabular}
\end{table}

\subsection{Computational Overhead}
\label{sec:cost}
We measure the computational overhead of TAPS during inference. 
All experiments were conducted on a GPU cluster consisting of 8 NVIDIA RTX A5000 GPUs and 4 NVIDIA RTX A40 GPUs. 
Evaluation runs were distributed across these GPUs, with the same hardware configuration used for the base model and TAPS whenever reporting paired runtime and memory comparisons. 
As shown in Table~\ref{tab:taps_cost}, compared with the base model, TAPS introduces only a marginal time overhead (+0.2\%) and no additional peak memory usage. 
These results indicate that TAPS is computationally lightweight in practice. The full evaluation, including baseline comparisons, ablations, and multi-sample generation, required approximately 800 GPU-hours in total.

\begin{table}[htbp]
\centering
\caption{Inference cost comparison between the base model and TAPS. Results are averaged over 30 trials on a fixed GSM8K prompt with 256-token generation at 128 denoising steps.}
\label{tab:taps_cost}
\begin{tabular}{lccccc}
\hline
Method & time/sample & time/step & peak mem & $\Delta$ Time & $\Delta$ Memory \\
\hline
Base (Normal) & 19.980 $\pm$ 0.010s & 156.09 ms & 16830 MB & -- & -- \\
TAPS          & 20.012 $\pm$ 0.006s & 156.34 ms & 16830 MB & +0.2\% & 0\% \\
\hline
\end{tabular}
\end{table}

\subsection{A Toy Experiment}
We design a toy experiment to provide an intuitive illustration of semantic branching and diversity evolution in diffusion language models. Using TraDo-8B as the backbone, we construct a semantically ambiguous prompt—\emph{``The mysterious prisoner looked at the guard and suddenly''}—which naturally admits multiple coherent but distinct continuations. We compare Standard DLM decoding with TAPS, where perturbations are injected during the early denoising stage (the first 30\% of steps) to encourage exploration of diverse semantic trajectories.

\begin{figure*}[h]
    \centering
    \includegraphics[width=\textwidth]{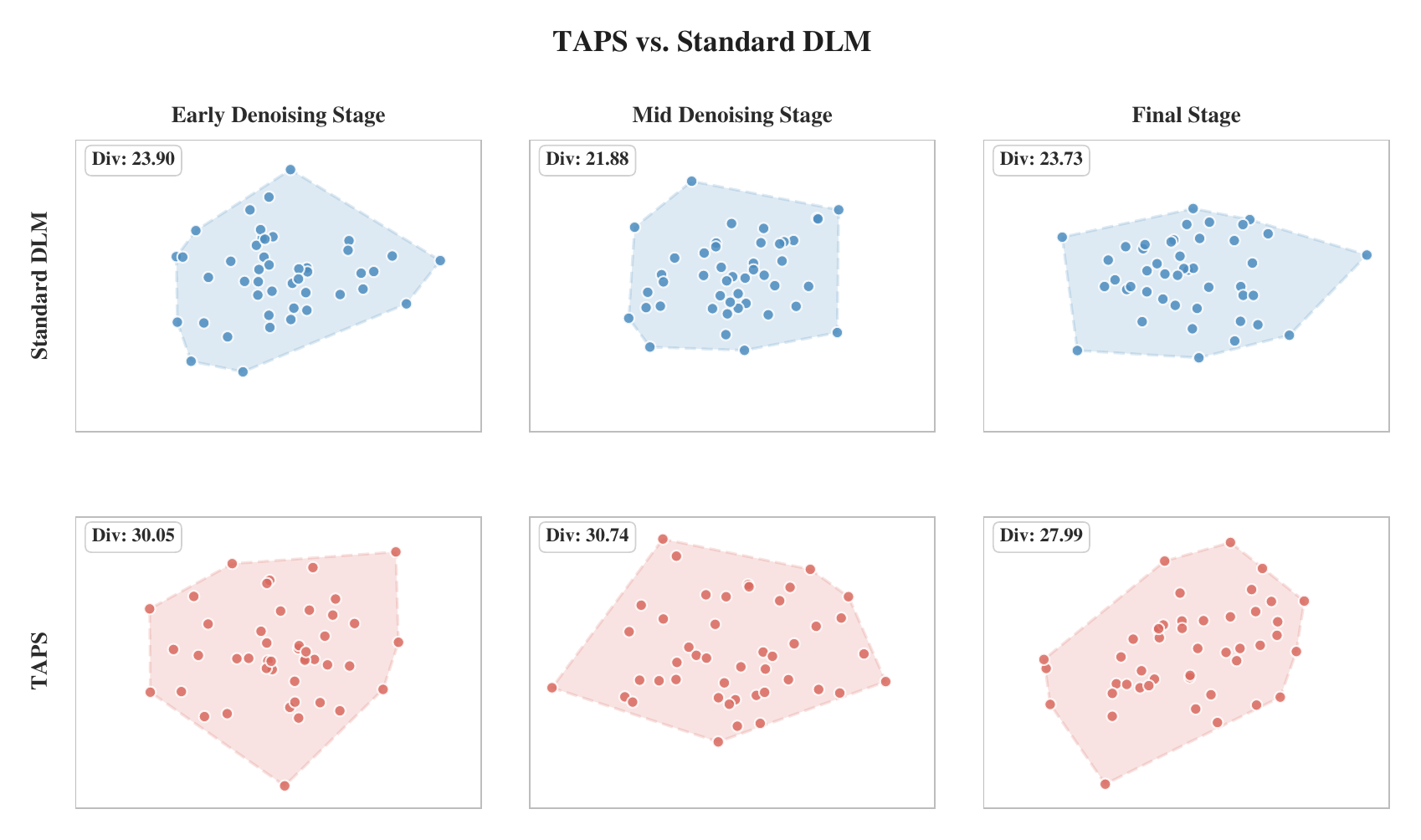}
    \vspace{-1mm}
    \caption{Toy experiment on semantic branching with TraDo-8B. We compare Standard DLM and TAPS by projecting Sent-BERT embeddings of generated samples into a shared 2D space via t-SNE at three denoising stages (early, mid, final). Convex hulls illustrate semantic coverage. TAPS maintains broader and more multimodal coverage across stages, while Standard DLM exhibits progressive semantic contraction. For the mid-stage snapshot, we retain all predicted tokens before re-masking when computing semantic representations.}
    \label{fig:semantic_fork}
    \vspace{-2mm}
\end{figure*}

To visualize the evolution of semantic diversity, we embed generated samples using Sent-BERT and project them into a shared two-dimensional space via t-SNE. We capture snapshots at three representative stages (early, middle, and final) and visualize semantic coverage using convex hulls, as shown in Figure~\ref{fig:semantic_fork}. For the intermediate stage, since DLMs predict all tokens at each denoising step followed by re-masking, we retain all predicted tokens when computing semantic representations. The results reveal a clear contrast: the Standard DLM exhibits progressive semantic contraction and mode collapse toward a single dominant region, whereas TAPS consistently maintains broader and more multimodal semantic coverage across all stages. This visualization highlights the importance of early-stage intervention in preserving semantic diversity and mitigating repetitive generation in diffusion language models.

\subsection{Token-level Mask-based Noise Injection}
\label{sec:token_noise}

In addition to embedding-level perturbation, we explore a token-level variant that injects noise by masking a subset of tokens during decoding. Due to architectural differences, this variant is evaluated only on LLaDA-8B-Instruct. Specifically, TraDo-8B-Instruct is adapted from autoregressive model weights and does not naturally support token-level masking within the diffusion decoding process.

For token-level perturbation, we randomly mask 5\% of the context tokens and apply the same time-annealed schedule as in TAPS, where the masking effect gradually decays within a predefined early denoising window. We conduct experiments at temperature $T=1.0$ on a writing prompt benchmark consisting of 100 samples, and compare token-level masking with embedding-level perturbation as well as the base decoding strategy.

As shown in Table~\ref{tab:token_mask}, both perturbation variants improve diversity-related metrics over the base model, indicating that TAPS can be instantiated using either discrete or continuous perturbation mechanisms. However, while token-level masking achieves comparable or higher gains in certain diversity metrics, it exhibits a larger degradation in quality-related scores. In contrast, embedding-level perturbation consistently maintains a more favorable balance between diversity and generation quality, suggesting that continuous perturbations offer better stability for diffusion language models in practice.

\begin{table*}[h]
\centering
\small
\setlength{\tabcolsep}{4.5pt}
\renewcommand{\arraystretch}{1.12}
\caption{Comparison between token-level mask-based perturbation and embedding-level perturbation on LLaDA-8B-Instruct at temperature $T=1.0$. Results are averaged over 100 writing prompts.}
\resizebox{\linewidth}{!}{%
\begin{tabular}{lcccccccc}
\toprule
& \multicolumn{4}{c}{\textbf{Diversity Metrics}} 
& \multicolumn{4}{c}{\textbf{Quality (GPT-4o)}} \\
\cmidrule(lr){2-5}\cmidrule(lr){6-9}
\textbf{Method} &
\textbf{IntraDistinct}~$\uparrow$ &
\textbf{Div-BLEU}~$\uparrow$ &
\textbf{Sent-BERT}~$\uparrow$ &
\textbf{EAD}~$\uparrow$ &
\textbf{Crea}~$\uparrow$ &
\textbf{Coh}~$\uparrow$ &
\textbf{WQ}~$\uparrow$ &
\textbf{Rel}~$\uparrow$ \\
\midrule
\multicolumn{9}{l}{\textbf{LLaDA-8B-Instruct}} \\
\midrule
Base Model
& 72.44 & 75.49 & 35.95 & 71.14 & 6.86 & \textbf{7.32} & 7.18 & 7.66 \\
TAPS (Token-level Mask)
& 71.14 & \textbf{76.69} & 36.12 & 72.43 & 6.81 & 7.18 & 6.95 & 7.60 \\
\rowcolor{black!6}
TAPS (Embedding-level)
& \textbf{72.68} & 76.60 & \textbf{36.51} & \textbf{72.68} & \textbf{6.90} & 7.23 & \textbf{7.25} & \textbf{7.73} \\
\bottomrule
\end{tabular}%
}

\label{tab:token_mask}
\end{table*}

\subsection{Results under Different Temperatures}
\label{sec:diff_temp}

Due to space constraints, we report results at temperature $T=1.0$ in the main paper. In this appendix, we further present results under additional temperature settings, $T=0.8$ and $T=1.5$, on NoveltyBench and Arena-Hard-Auto. We limit our evaluation to these three values, as diffusion language models tend to produce unstable or degenerate outputs when the temperature falls outside this range. Experimental results demonstrate that, across different temperatures, our method consistently improves diversity-related metrics while maintaining comparable generation quality.

\subsubsection{Arena-Hard-Auto}

For Arena-Hard-Auto, the main paper reports results on the full creative writing subset. In this appendix, we additionally evaluate on a randomly sampled subset of 50 creative writing prompts, where each prompt is generated 8 times to facilitate diversity evaluation. Results under different temperatures are reported in Table~\ref{tab:Area_quality_div_temp0.8} ($T=0.8$) and Table~\ref{tab:Area_quality_div_temp1.5} ($T=1.5$).

At lower temperature ($T=0.8$), our method consistently outperforms baseline decoding strategies on most diversity metrics across both backbones. Notably, TAPS also achieves the best performance on multiple quality-related metrics, indicating that time-annealed perturbation can effectively enhance diversity without sacrificing generation quality in conservative decoding regimes. At a higher temperature ($T=1.5$), we observe distinct behaviors between the two backbones. For LLaDA-8B-Instruct, a fully non-autoregressive diffusion model, TAPS remains robust under increased sampling stochasticity and continues to achieve strong diversity gains while maintaining competitive quality. In contrast, for TraDo-8B-Instruct, which retains certain semi-autoregressive characteristics, high temperature amplifies the inherent randomness of autoregressive decoding, leading to noticeable quality degradation for several baseline methods. Under this setting, Top-$k$ sampling achieves the strongest overall quality performance on TraDo-8B-Instruct. This behavior is expected, as Top-$k$ explicitly filters out low-probability tokens and thus provides a stronger constraint on local token selection. Given TraDo’s partial autoregressive nature, such token-level filtering is particularly effective at mitigating error accumulation under high-temperature decoding.

Overall, these results suggest that TAPS provides stable and effective diversity control across temperature regimes, with its robustness being particularly pronounced for diffusion-style backbones. The observed differences further highlight how the degree of autoregressive structure influences a model’s tolerance to high-temperature stochasticity.

\begin{table*}[h]
\centering
\small
\setlength{\tabcolsep}{4.5pt}
\renewcommand{\arraystretch}{1.12}
\caption{Diversity and multi-aspect quality comparison across two backbones in Arena-Hard-Auto (T=0.8).}
\resizebox{\linewidth}{!}{%
\begin{tabular}{lcccccccc}
\toprule
& \multicolumn{4}{c}{\textbf{Diversity metrics}} & \multicolumn{4}{c}{\textbf{Quality (GPT-4o)}} \\
\cmidrule(lr){2-5}\cmidrule(lr){6-9}
\textbf{Backbone / Method} &
\textbf{IntraDistinct}~$\uparrow$ &
\textbf{Div-BLEU}~$\uparrow$ &
\textbf{Sent-BERT}~$\uparrow$ &
\textbf{EAD}~$\uparrow$ &
\textbf{Crea}~$\uparrow$ &
\textbf{Coh}~$\uparrow$ &
\textbf{WQ}~$\uparrow$ &
\textbf{Rel}~$\uparrow$ \\
\midrule

\multicolumn{9}{l}{\textbf{LLaDA-8B-Instruct}}\\
\midrule
Base (Normal)          & 75.35 & 40.81 & 20.66 & 48.22 & 4.85 & 5.02 & 5.20 & 5.08 \\
Top-$p$                & 76.82 & 38.85 & 19.16 & 45.48 & 4.70 & 4.82 & 4.87 & 4.97 \\
Top-$k$                & 75.07 & \textbf{44.98} & 23.27 & \textbf{50.23} & 4.94 & 5.04 & 4.97 & \textbf{5.25} \\
Min-$p$                & 77.15 & 33.09 & 14.95 & 40.90 & 4.12 & 4.36 & 4.43 & 4.40 \\
EDT                    & 50.50 & 27.44 & 10.08 & 21.80 & 4.20 & 4.23 & 4.31 & 4.25 \\
Diverse Prompt         & 69.98 & 10.87 & 10.31 & 27.97 & 4.75 & 4.72 & 4.77 & 4.81 \\
\rowcolor{black!6}
\textbf{TAPS}   & \textbf{78.47} & 43.57 & \textbf{24.63} & 49.35
                       & \textbf{4.96} & \textbf{5.07} & \textbf{5.23} & 5.21 \\
\midrule

\multicolumn{9}{l}{\textbf{TraDo-8B-Instruct}}\\
\midrule
Base (Normal)          & 81.80 & 40.45 & 20.29 & 49.72 & 5.83 & 5.12 & 5.07 & 6.02 \\
Top-$p$                & 81.63 & 38.22 & 19.90 & 48.40 & 5.80 & 5.19 & 5.16 & 6.03 \\
Top-$k$                & 81.64 & 40.36 & 20.25 & 49.54 & 5.87 & \textbf{5.20} & 5.12 & 6.02 \\
Min-$p$                & 80.51 & 37.39 & 19.65 & 47.45 & 5.65 & 5.05 & 5.06 & 5.90 \\
EDT                    & 80.50 & 27.44 & 10.08 & 21.80 & 3.40 & 2.09 & 2.28 & 2.31 \\
Diverse Prompt         & 69.83 & 17.06 & 6.30 & 19.37 & 3.55 & 2.62 & 2.79 & 2.77 \\
\rowcolor{black!6}
\textbf{TAPS}   & \textbf{82.47} & \textbf{40.77} & \textbf{20.36} & \textbf{50.30}
                       & \textbf{5.90} & 5.19 & \textbf{5.09} & \textbf{6.07} \\
\bottomrule
\end{tabular}%
}

\label{tab:Area_quality_div_temp0.8}
\end{table*}

\begin{table*}[h]
\centering
\small
\setlength{\tabcolsep}{4.5pt}
\renewcommand{\arraystretch}{1.12}
\caption{Diversity and multi-aspect quality comparison across two backbones in Arena-Hard-Auto (T=1.5).}
\resizebox{\linewidth}{!}{%
\begin{tabular}{lcccccccc}
\toprule
& \multicolumn{4}{c}{\textbf{Diversity metrics}} & \multicolumn{4}{c}{\textbf{Quality (GPT-4o)}} \\
\cmidrule(lr){2-5}\cmidrule(lr){6-9}
\textbf{Backbone / Method} &
\textbf{IntraDistinct}~$\uparrow$ &
\textbf{Div-BLEU}~$\uparrow$ &
\textbf{Sent-BERT}~$\uparrow$ &
\textbf{EAD}~$\uparrow$ &
\textbf{Crea}~$\uparrow$ &
\textbf{Coh}~$\uparrow$ &
\textbf{WQ}~$\uparrow$ &
\textbf{Rel}~$\uparrow$ \\
\midrule

\multicolumn{9}{l}{\textbf{LLaDA-8B-Instruct}}\\
\midrule
Base (Normal)          & 83.33 & 54.02 & \textbf{32.44} & 64.50 & 4.08 & 3.75 & 3.73 & 4.14 \\
Top-$p$                & 79.76 & 49.30 & 27.31 & 54.40 & 4.60 & 4.61 & 4.66 & 4.78 \\
Top-$k$                & 75.98 & 47.34 & 28.60 & 49.14 & 4.37 & 4.65 & 4.79 & \textbf{4.84} \\
Min-$p$                & 80.27 & 37.47 & 21.35 & 44.06 & 4.21 & 4.35 & 4.63 & 4.42 \\
EDT                    & 79.12 & 42.44 & 22.79 & 41.10 & 4.13 & 4.21 & 4.48 & 4.29 \\
Diverse Prompt         & 75.43 & 29.38 & 18.16 & 41.21 & 4.48 & 3.75 & 3.76 & 4.48 \\
\rowcolor{black!6}
\textbf{TAPS}   & \textbf{84.29} & \textbf{55.58} & 32.14 & \textbf{65.80}
                       & \textbf{4.61} & \textbf{4.72} & \textbf{4.93} & 4.77 \\
\midrule

\multicolumn{9}{l}{\textbf{TraDo-8B-Instruct}}\\
\midrule
Base (Normal)          & 87.51 & 58.32 & 23.18 & 60.47 & 3.27 & 1.85 & 1.77 & 2.57 \\
Top-$p$                & 87.33 & 57.38 & 23.21 & 61.21 & 3.52 & 2.08 & 2.23 & 3.04 \\
Top-$k$                & 83.05 & 44.26 & 12.25 & 52.21 & \textbf{5.00} & \textbf{4.31} & \textbf{4.28} & \textbf{4.97} \\
Min-$p$                & 78.45 & 37.78 & 9.30 & 46.31 & 4.82 & 4.23 & 4.19 & 4.80 \\
EDT                    & 82.12 & 29.20 & 12.09 & 24.75 & 2.51 & 1.48 & 1.66 & 1.61 \\
Diverse Prompt         & 70.93 & 18.11 & 9.38 & 22.78 & 2.62 & 1.69 & 1.83 & 1.83 \\
\rowcolor{black!6}
\textbf{TAPS}   & \textbf{88.30} & \textbf{59.58} & \textbf{24.04} & \textbf{61.80}
                       & 4.26 & 3.53 & 3.20 & 3.72 \\
\bottomrule
\end{tabular}%
}
\label{tab:Area_quality_div_temp1.5}
\end{table*}

\subsubsection{NoveltyBench}

For Novelty-Bench, we additionally evaluate on both the curated and wildchat subsets by randomly sampling 50 prompts from each subset. For each prompt, we generate 10 samples and report the average results. Table~\ref{tab:div_novelty_temp0.8_1.5} summarizes the diversity metrics under different temperature settings.

\begin{figure}[h]
    \centering
    \includegraphics[width=0.7\linewidth]{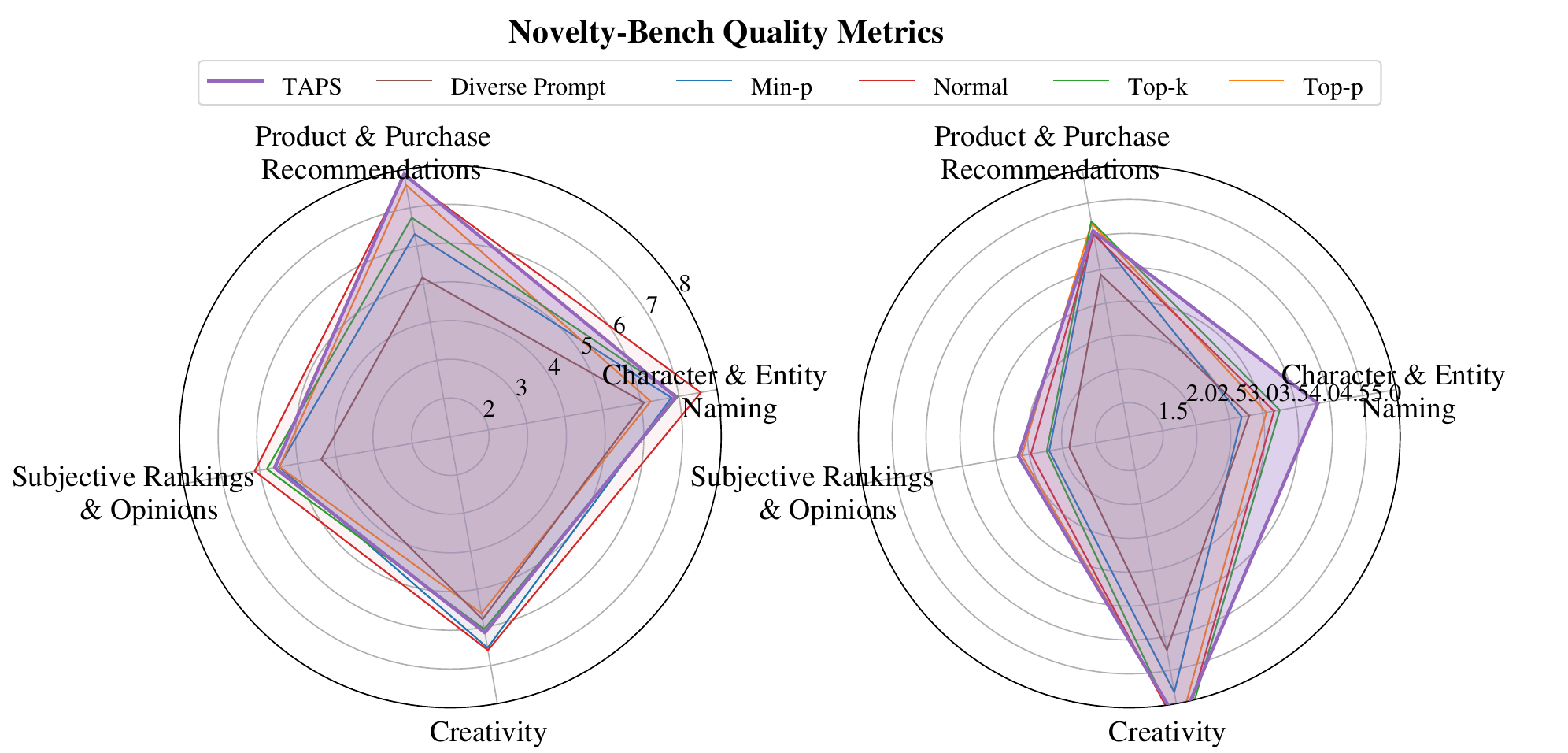}
    \caption{Multi-aspect quality comparison on Novelty-Bench at temperature $T=0.8$. We evaluate generation quality across four dimensions: Product \& Purchase Recommendations, Character \& Entity Naming, Subjective Rankings \& Opinions, and Creativity. TAPS maintains strong and balanced performance across all aspects, while some baseline methods, such as Diverse Prompt, exhibit noticeable quality degradation despite increased diversity.}
    \label{fig:radar_0.8}
\end{figure}

At lower temperature ($T=0.8$), our method consistently achieves the best or near-best performance across almost all diversity-related metrics on both backbones, demonstrating its effectiveness in promoting diverse generation under conservative decoding regimes. Importantly, these diversity gains do not come at the expense of generation quality: as shown in Figure~\ref{fig:radar_0.8}, TAPS maintains competitive or superior quality scores across multiple categories, including creativity and subjective preference, compared to baseline decoding strategies. In contrast, the Diverse Prompt baseline exhibits noticeable quality degradation across several aspects, suggesting that aggressively encouraging diversity through prompt manipulation can negatively impact overall generation quality.

At higher temperature ($T=1.5$), we observe a different behavior. For the base model, this temperature already approaches the upper limit of stable generation, leading to a higher likelihood of degenerate or incoherent outputs. Under this setting, several methods exhibit a sharp increase in diversity metrics, including TAPS and Diverse Prompt, but this is accompanied by substantial degradation in generation quality. In contrast, Top-$p$ and Top-$k$ sampling retain relatively strong quality performance at high temperature, which aligns with their intrinsic filtering mechanisms that suppress low-probability or implausible tokens. These results suggest that while high temperature can artificially inflate diversity scores, it may do so by sacrificing output quality, and highlight the importance of balanced diversity control under realistic decoding regimes.

\begin{table*}[t]
\centering
\small
\setlength{\tabcolsep}{3pt}
\renewcommand{\arraystretch}{1.12}
\caption{Diversity comparison across two backbones on Novelty-Bench under different temperatures. Left: $T=0.8$. Right: $T=1.5$.}
\begin{minipage}[t]{0.49\textwidth}
\centering

\resizebox{\linewidth}{!}{
\begin{tabular}{lcccc}
\toprule
& \multicolumn{4}{c}{\textbf{Diversity metrics}} \\
\cmidrule(lr){2-5}
\textbf{Backbone / Method} &
\textbf{IntraDistinct}~$\uparrow$ &
\textbf{Div-BLEU}~$\uparrow$ &
\textbf{Sent-BERT}~$\uparrow$ &
\textbf{EAD}~$\uparrow$ \\
\midrule

\multicolumn{5}{l}{\textbf{LLaDA-8B-Instruct}}\\
\midrule
Base (Normal)          & 80.13 & 46.64 & 34.57 & 54.05 \\
Top-$p$                & 77.91 & 47.24 & \textbf{36.75} & 54.24 \\
Top-$k$                & 81.52 & 46.53 & 34.49 & 53.97 \\
Min-$p$                & 81.39 & 40.82 & 32.88 & 52.18 \\
EDT                    & -- & -- & -- & -- \\
Diverse Prompt         & 77.86 & 36.09 & 27.07 & 45.64 \\
\rowcolor{black!6}
\textbf{TAPS}   & \textbf{82.27} & \textbf{47.61} & 36.64 & \textbf{56.14} \\
\midrule

\multicolumn{5}{l}{\textbf{TraDo-8B-Instruct}}\\
\midrule
Base (Normal)          & 84.79 & 72.60 & 29.33 & 74.88 \\
Top-$p$                & 84.17 & 72.82 & 28.81 & 74.88 \\
Top-$k$                & 84.33 & 72.23 & 28.96 & 74.75 \\
Min-$p$                & 84.28 & 72.12 & 29.18 & 73.95 \\
EDT                    & -- & -- & -- & -- \\
Diverse Prompt         & 84.02 & 68.28 & 30.21 & 72.84 \\
\rowcolor{black!6}
\textbf{TAPS}   & \textbf{85.60} & \textbf{73.91} & \textbf{31.20} & \textbf{75.72} \\
\bottomrule
\end{tabular}}
\end{minipage}
\hfill
\begin{minipage}[t]{0.49\textwidth}
\centering

\resizebox{\linewidth}{!}{
\begin{tabular}{lcccc}
\toprule
& \multicolumn{4}{c}{\textbf{Diversity metrics}} \\
\cmidrule(lr){2-5}
\textbf{Backbone / Method} &
\textbf{IntraDistinct}~$\uparrow$ &
\textbf{Div-BLEU}~$\uparrow$ &
\textbf{Sent-BERT}~$\uparrow$ &
\textbf{EAD}~$\uparrow$ \\
\midrule

\multicolumn{5}{l}{\textbf{LLaDA-8B-Instruct}}\\
\midrule
Base (Normal)          & 79.01 & 62.33 & 30.89 & 61.18 \\
Top-$p$                & 84.76 & 77.99 & 41.65 & 77.14 \\
Top-$k$                & 82.70 & 72.36 & 36.76 & 72.49 \\
Min-$p$                & 82.03 & 65.81 & 33.02 & 65.61 \\
EDT                    & -- & -- & -- & -- \\
Diverse Prompt         & 86.10 & 80.45 & 41.52 & 79.23 \\
\rowcolor{black!6}
\textbf{TAPS}   & \textbf{88.33} & \textbf{88.84} & \textbf{49.12} & \textbf{84.11} \\
\midrule

\multicolumn{5}{l}{\textbf{TraDo-8B-Instruct}}\\
\midrule
Base (Normal)          & 84.79 & 72.60 & 29.33 & 74.88 \\
Top-$p$                & 87.53 & 85.58 & 36.70 & 82.77 \\
Top-$k$                & 87.30 & 89.43 & 38.88 & 83.81 \\
Min-$p$                & 84.97 & 81.67 & 32.69 & 79.06 \\
EDT                    & -- & -- & -- & -- \\
Diverse Prompt         & \textbf{98.02} & 96.90 & 55.21 & \textbf{104.57} \\
\rowcolor{black!6}
\textbf{TAPS}   & 94.45 & \textbf{98.69} & \textbf{56.08} & 94.15 \\
\bottomrule
\end{tabular}}
\end{minipage}
\label{tab:div_novelty_temp0.8_1.5}
\end{table*}

\subsection{Preference Model Evaluation}
\label{sec:pref_model}

To complement the automatic quality metrics, we conduct an LLM-based pairwise preference evaluation following the comparison protocol used in Arena-Hard-Auto and WritingPrompts. 
For each prompt, the evaluator is shown two anonymized responses generated by different methods and asked to select the one with better overall quality, considering helpfulness, coherence, relevance, and creativity. 
To reduce positional bias, the response order is randomized across comparisons. 
We use GPT-4o and Gemini-2 as independent preference models and report the average win rate against the baseline decoding method.

As shown in Table~\ref{tab:arena_hard_results}, most decoding strategies, including Min-$p$, Top-$p$, Top-$k$, the Base Model, and TAPS, achieve comparable preference scores around 50\%, indicating similar overall quality under LLM-based preference evaluation. 
TAPS obtains the second-highest win rate (51.7\%), closely following Min-$p$ (52.6\%), and their confidence intervals largely overlap, suggesting no statistically significant difference. 
In contrast, EDT and Diverse Prompt perform substantially worse, indicating that aggressively promoting diversity can harm preference-aligned quality. 
These results show that TAPS improves diversity while maintaining competitive preference-level quality.

\begin{table}[htbp]
\centering
\caption{Win-rate statistics from LLM-based pairwise preference evaluation on Arena-Hard-v2.0 (Creative Writing) and WritingPrompts.}
\label{tab:arena_hard_results}
\begin{tabular}{lcc}
\toprule
\textbf{Method} & \textbf{Scores (\%)} & \textbf{CI (\%)} \\
\midrule
Min-$p$ & 52.6 & (-10.7 / +7.9) \\
\rowcolor{black!6}
TAPS & 51.7 & (-7.2 / +9.1) \\
Top-$p$ & 50.5 & (-8.6 / +6.9) \\
Base Model & 50.0 & (-0.0 / +0.0) \\
Top-$k$ & 49.3 & (-9.8 / +8.8) \\
EDT & 22.3 & (-7.1 / +8.2) \\
diverse prompt & 19.2 & (-6.7 / +9.5) \\
\bottomrule
\end{tabular}
\end{table}

\section{Additional Robustness Study}
\label{sec:add_ablation}

\subsection{Studied Hyperparameters}

In the robustness study, we vary the following hyperparameters of TAPS while keeping all others fixed to their default values. We consider (i) the noise scale($\sigma$), (ii) noise injection time windows $t$ defined by the start and end denoising steps, which are treated jointly as they together determine the duration of perturbation, and (iii) the noise annealing strategy, including cosine and linear decay. Unless otherwise specified, all results are reported using the same decoding configuration as in the main experiments.

\subsection{Budget-Matched Perturbation Timing}
\label{sec:budget_matched_timing}

To separate the effect of perturbation timing from the total amount of injected noise, we compare four schedules under the same accumulated perturbation budget in Table~\ref{tab:budget_matched_timing}, $\delta_0=21.25$. The schedules apply perturbation early only, late only, throughout the trajectory, or with the annealed TAPS schedule. Early-only perturbation achieves the highest semantic diversity but reduces quality. Annealed TAPS retains most of this diversity while achieving the highest quality. Late-only and full-trajectory perturbations yield weaker semantic diversity under the same budget, supporting the choice to concentrate exploration early and reduce perturbations before refinement.

\begin{table}[htbp]
\centering
\small
\caption{Perturbation timing under a matched accumulated budget ($\delta_0=21.25$).}
\label{tab:budget_matched_timing}
\begin{tabular}{lcc}
\toprule
\textbf{Schedule} & \textbf{Sent-BERT} $\uparrow$ & \textbf{Quality} $\uparrow$ \\
\midrule
Annealed (TAPS) & 36.79 & \textbf{6.44} \\
Early only & \textbf{37.60} & 6.20 \\
Late only & 36.31 & 6.24 \\
Full trajectory & 34.87 & 6.24 \\
\bottomrule
\end{tabular}
\end{table}

\subsection{Robustness w.r.t. Projection Operators}
\label{sec:projection_operators}

We compare three projection operators on a diagnostic LLaDA-8B-Instruct/WritingPrompts subset in Table~\ref{tab:projection_operators}, holding the perturbation, annealing, rescaling, and mixing steps fixed.

\begin{table}[htbp]
\centering
\small
\caption{Diagnostic comparison of projection operators on LLaDA-8B-Instruct and a WritingPrompts subset.}
\label{tab:projection_operators}
\resizebox{\linewidth}{!}{%
\begin{tabular}{lccccc}
\toprule
\textbf{Projection} & \textbf{IntraDistinct} $\uparrow$
& \textbf{Div-BLEU} $\uparrow$ & \textbf{Sent-BERT} $\uparrow$
& \textbf{EAD} $\uparrow$ & \textbf{Quality} $\uparrow$ \\
\midrule
Global norm & 56.97 & 67.56 & 35.13 & 59.31 & 5.83 \\
Token norm (TAPS) & 57.38 & 65.75 & \textbf{35.28} & 59.21 & \textbf{6.05} \\
Local subspace & \textbf{58.13} & \textbf{67.67} & 34.49 & \textbf{59.77} & 5.98 \\
\bottomrule
\end{tabular}%
}
\end{table}

Token-wise norm projection achieves the highest quality and semantic diversity in this diagnostic comparison, whereas prompt-local subspace projection yields higher lexical diversity. Thus, token-wise projection is a simple, effective choice, although the preferred operator may depend on the desired quality--diversity trade-off.

\subsection{Robustness w.r.t. Noise Distribution}
\label{sec:noise_distribution}

As shown in Table~\ref{tab:noise_distribution}, we compare Gaussian, uniform, and Laplace perturbations within the same TAPS pipeline in a small-scale diagnostic experiment.

\begin{table}[htpb]
\centering
\small
\caption{Diagnostic comparison of perturbation noise distributions. The high Sent-BERT score under Laplace noise should not be interpreted as useful diversity because outputs are largely degenerate.}
\label{tab:noise_distribution}
\begin{tabular}{lccc}
\toprule
\textbf{Noise} & \textbf{IntraDistinct} $\uparrow$
& \textbf{Sent-BERT} $\uparrow$ & \textbf{Avg. Quality} $\uparrow$ \\
\midrule
Uniform & 61.5 & 39.9 & 6.39 \\
Gaussian & \textbf{62.1} & 40.1 & \textbf{6.45} \\
Laplace & 4.5 & \textbf{42.1} & 1.43 \\
\bottomrule
\end{tabular}
\end{table}

Gaussian and uniform noise give similar diversity and quality, whereas Laplace noise produces largely degenerate outputs and a substantial quality drop. Its elevated Sent-BERT score therefore does not represent useful semantic variation. Heavy-tailed coordinate outliers are a possible explanation, but this diagnostic alone does not establish the cause.

\subsection{Robustness w.r.t. Annealing Strategy}
\label{sec:noise_strategy}

Overall, as shown in Table~\ref{tab:anneal_strategy}, we observe that TAPS exhibits similar performance under both cosine and linear annealing schedules. This suggests that the method does not critically rely on a specific annealing function, provided that the injected noise decays over time. More specifically, for both LLaDA-8B-Instruct and TraDo-8B-Instruct, introducing TAPS with either linear or cosine annealing consistently improves diversity-related metrics over the base decoding strategy. While cosine annealing achieves slightly higher scores in most diversity metrics, the overall performance gap between the two schedules remains small, and both variants exhibit comparable quality scores across creativity, coherence, writing quality, and relevance. These results indicate that the effectiveness of TAPS is robust to the choice of annealing strategy, and that monotonic noise decay—rather than a specific functional form—is the key factor in promoting diverse generation.

\begin{table}[!h]
\centering
\small
\setlength{\tabcolsep}{4pt}
\renewcommand{\arraystretch}{1.12}
\caption{Robustness study w.r.t.\ annealing strategy. We compare cosine versus linear noise decay on two backbones while keeping other hyperparameters fixed.}
\resizebox{\linewidth}{!}{
\begin{tabular}{lcccccccc}
\toprule
& \multicolumn{4}{c}{\textbf{Diversity metrics}} 
& \multicolumn{4}{c}{\textbf{Quality (GPT-4o)}} \\
\cmidrule(lr){2-5}\cmidrule(lr){6-9}
\textbf{Backbone / Method} &
\textbf{IntraDistinct}~$\uparrow$ &
\textbf{Div-BLEU}~$\uparrow$ &
\textbf{Sent-BERT}~$\uparrow$ &
\textbf{EAD}~$\uparrow$ &
\textbf{Crea}~$\uparrow$ &
\textbf{Coh}~$\uparrow$ &
\textbf{WQ}~$\uparrow$ &
\textbf{Rel}~$\uparrow$ \\
\midrule

\multicolumn{9}{l}{\textbf{LLaDA-8B-Instruct}}\\
\midrule
Base (Normal)            & 57.82 & 65.11 & 33.53 & 57.60 & 6.25 & 5.66 & 6.04 & 6.76 \\
TAPS (Linear Anneal)      & 59.06 & 66.32 & 34.82 & 58.44 & 6.30 & 5.74 & 6.03 & 6.73 \\
\rowcolor{black!6}
TAPS (Cosine Anneal)      & 59.73 & 66.38 & 34.41 & 59.20 & 6.31 & 5.79 & 6.11 & 6.65 \\
\midrule

\multicolumn{9}{l}{\textbf{TraDo-8B-Instruct}}\\
\midrule
Base (Normal)            & 76.02 & 89.60 & 56.63 & 77.65 & 5.03 & 5.17 & 5.06 & 5.92 \\
TAPS (Linear Anneal)      & 76.88 & 89.87 & 57.09 & 78.04 & 5.08 & 5.17 & 5.10 & 5.81 \\
\rowcolor{black!6}
TAPS (Cosine Anneal)      & 77.07 & 89.93 & 57.40 & 78.22  & 5.07 & 5.19 & 5.11 & 5.83 \\
\bottomrule
\end{tabular}%
}
\label{tab:anneal_strategy}
\end{table}

\subsection{Robustness w.r.t. Backbone}
\label{sec:robustness_backbone}

\subsubsection{Opening Generation}
\label{sec:add_backbone_open}
To further study the robustness of our method across different backbones, we additionally evaluate TAPS on two diffusion language models: \textsc{Dream-v0-Instruct-7B} and \textsc{SDAR-VL-Instruct-8B}. Specifically, we conduct experiments on two benchmarks, NoveltyBench and Arena-Hard-Auto, using these models as alternative backbones. The results are reported in Tables~\ref{tab:robustness_dream} and~\ref{tab:robustness_sadr}. Overall, TAPS shows consistent improvements across both benchmarks and both backbones, indicating that its effectiveness generalizes beyond the models used in the main experiments.

\begin{table}[htbp]
\centering
\caption{Robustness study on the \textsc{Dream} backbone. We evaluate different decoding methods on \textsc{NoveltyBench} and \textsc{Arena-Hard-Auto}.}
\label{tab:robustness_dream}
\begin{tabular}{llcccc}
\hline
 & & \multicolumn{4}{c}{Diversity metrics} \\
\cline{3-6}
Benchmark & Method & IntraDistinct$\uparrow$ & Div-Bleu$\uparrow$ & Sent-BERT$\uparrow$ & EAD$\uparrow$ \\
\hline
\multirow{5}{*}{NoveltyBench}
& Base (Normal) & \textbf{84.20} & 69.80 & 46.43 & 78.84 \\
& Top-$k$       & 84.16 & 67.85 & 45.74 & 78.35 \\
& Top-$p$       & 83.04 & 62.07 & 44.75 & 74.94 \\
& Min-$p$       & 83.06 & 62.57 & 45.77 & 75.50 \\
\rowcolor{black!6}& TAPS          & 83.83 & \textbf{74.92} & \textbf{54.48} & \textbf{82.46} \\
\hline
\multirow{5}{*}{Arena-Hard-Auto}
& Base (Normal) & 75.72 & 89.02 & 49.04 & 72.92 \\
& Top-$k$       & 73.37 & 87.28 & 45.24 & 71.06 \\
& Top-$p$       & 75.14 & 86.64 & 44.91 & 72.47 \\
& Min-$p$       & 73.92 & 86.98 & 44.82 & 70.61 \\
\rowcolor{black!6}& TAPS          & \textbf{77.73} & \textbf{95.64} & \textbf{60.87} & \textbf{75.37} \\
\hline
\end{tabular}
\end{table}

\begin{table}[htbp]
\centering
\caption{Robustness study on the \textsc{SDAR} backbone. We evaluate different decoding methods on \textsc{NoveltyBench} and \textsc{Arena-Hard-Auto}.}
\label{tab:robustness_sadr}
\begin{tabular}{llcccc}
\hline
Benchmark & Method & IntraDistinct $\uparrow$ & EAD $\uparrow$ & Sent-BERT $\uparrow$ & Div-BLEU $\uparrow$ \\
\hline
\multirow{5}{*}{NoveltyBench}
& Base (Normal) & 84.20 & 53.78 & 26.16 & 39.34 \\
& Top-$k$       & 84.59 & 54.47 & 25.97 & 39.01 \\
& Top-$p$       & 84.49 & 49.24 & 23.95 & 34.43 \\
& Min-$p$       & 84.66 & 49.47 & 23.05 & 33.38 \\
\rowcolor{black!6}& TAPS          & \textbf{84.48} & \textbf{55.01} & \textbf{28.10} & \textbf{40.92} \\
\hline
\multirow{5}{*}{Arena-Hard-Auto}
& Base (Normal) & 67.14 & 60.56 & 30.44 & 75.80 \\
& Top-$k$       & 65.61 & 58.60 & 29.65 & 74.63 \\
& Top-$p$       & 65.56 & 57.55 & 27.94 & 73.38 \\
& Min-$p$       & 64.00 & 55.99 & 28.07 & 72.59 \\
\rowcolor{black!6}& TAPS          & \textbf{68.37} & \textbf{62.20} & \textbf{33.22} & \textbf{77.58} \\
\hline
\end{tabular}
\end{table}

\subsubsection{Reasoning Ability}
\label{sec:add_backbone_reason}

We further examine whether TAPS affects the reasoning capability of different backbones on GSM8K. As shown in Tables~\ref{tab:gsm8k_additional_backbones}, TAPS consistently maintains or improves reasoning performance compared with the base model across both SDAR-8B-Chat and Dream-7B-Instruct. On SDAR-8B-Chat, TAPS achieves the strongest mean and majority-voting performance while keeping Pass@5 comparable to the best baseline. On Dream-7B-Instruct, although nucleus sampling performs best, TAPS still improves over the base model across all metrics. These results suggest that TAPS does not compromise reasoning ability when transferred to different diffusion language model backbones, and can preserve or even enhance reasoning performance in practice.

\begin{table}[!t]
\centering
\small
\setlength{\tabcolsep}{5pt}
\renewcommand{\arraystretch}{1.08}
\caption{GSM8K reasoning performance on additional diffusion backbones. We report mean accuracy with 95\% confidence intervals, Pass@5, and majority-vote accuracy.}
\label{tab:gsm8k_additional_backbones}
\resizebox{\linewidth}{!}{%
\begin{tabular}{llccc}
\toprule
\textbf{Backbone} & \textbf{Method} & \textbf{Mean $\pm$ 95\% CI} & \textbf{Pass@5} & \textbf{Majority Voting} \\
\midrule

\multirow{5}{*}{TraDo-8B-Instruct}
& Base model & 92.74 $\pm$ 2.17 & 97.17 & 94.86 \\
& Top-$p$    & \textbf{93.09 $\pm$ 2.09} & \underline{97.43} & \textbf{95.63} \\
& Top-$k$    & 92.91 $\pm$ 2.16 & 97.18 & 94.91 \\
& Min-$p$    & 92.80 $\pm$ 2.17 & 97.32 & 95.11 \\
\rowcolor{black!6}
& \textbf{TAPS} & \underline{92.95 $\pm$ 2.15} & \textbf{97.71} & \underline{95.73} \\

\midrule
\multirow{5}{*}{SDAR-8B-Chat}
& Base model & 81.82 $\pm$ 1.67 & \textbf{93.48} & 86.50 \\
& Top-$p$    & 82.02 $\pm$ 1.74 & 91.51 & 86.35 \\
& Top-$k$    & 81.82 $\pm$ 1.67 & \textbf{93.48} & 86.50 \\
& Min-$p$    & \underline{82.30 $\pm$ 1.71} & 91.58 & \underline{87.11} \\
\rowcolor{black!6}
& \textbf{TAPS} & \textbf{83.55 $\pm$ 1.61} & \underline{93.40} & \textbf{88.48} \\

\midrule
\multirow{5}{*}{Dream-7B-Instruct}
& Base model & 25.94 $\pm$ 1.60 & 58.38 & 35.94 \\
& Top-$p$    & \textbf{31.77 $\pm$ 1.73} & \textbf{63.84} & \textbf{43.97} \\
& Top-$k$    & 26.43 $\pm$ 1.60 & 58.30 & 36.32 \\
& Min-$p$    & \underline{30.13 $\pm$ 1.68} & \underline{63.53} & \underline{40.71} \\
\rowcolor{black!6}
& \textbf{TAPS} & 29.13 $\pm$ 1.64 & 61.47 & 40.36 \\
\bottomrule
\end{tabular}%
}
\end{table}

\section{Case Study}

\subsection{Examples on GSM8K Reasoning}
\label{sec:gsm8k_examples}

In this section, we present qualitative comparisons between our method and the base model on GSM8K problems with different levels of difficulty. We focus on two representative cases to illustrate how diversity-enhancing perturbations interact with mathematical reasoning under repeated sampling.

For an easier problem, we show that our method substantially increases the diversity of generated solutions while preserving correct reasoning, producing multiple valid reasoning paths that differ in their intermediate steps and descriptions. 
For a more challenging problem, where the solution space contains multiple uncertain or misleading paths, we demonstrate that injecting noise encourages broader exploration and helps prevent repeated sampling from collapsing into the same erroneous reasoning pattern. 
Together, these examples provide concrete insight into how controlled exploration enables diverse reasoning paths and benefits aggregation-based evaluation.

\paragraph{Example 1.}
Lloyd's chickens produce 252 eggs per day. He sells eggs at \$2 per dozen. How much money does he make in one week?

\medskip

\begin{tcolorbox}[
  breakable,
  enhanced,
  colback=blue!5,
  colframe=blue!40,
  title={Example 1: Comparison of 10 reasoning trajectories under different decoding methods},
  fonttitle=\bfseries,
  boxrule=0.6pt
]
\small

\textbf{TAPS (accuracy 100\%)}\\
(1) Lloyd's chickens produce 252 eggs per day. Since one dozen equals 12 eggs, 252 / 12 = 21 dozen eggs per day. He sells each dozen for \$2, so he makes 21 × \$2 = \$42 per day. In a week (7 days), he makes \$42 × 7 = \$294. The answer is \$294.

(2) 252 eggs per day equals 252 / 12 = 21 dozens of eggs. If Lloyd sells each dozen for \$2, then he makes 21 × \$2 = \$42 per day. In a week, Lloyd makes \$42 × 7 = \$294. The answer is \$294.

(3) Lloyd's chickens produce 252 eggs per day. Since there are 12 eggs in a dozen, 252 / 12 = 21 dozens per day. He sells them for \$2 per dozen, so he makes \$42 per day. Over 7 days, he earns \$294. The answer is \$294.

(4) 252 eggs per day means 252 × 7 = 1764 eggs per week. Dividing by 12 eggs per dozen gives 147 dozens per week. At \$2 per dozen, Lloyd earns \$294 per week. The answer is \$294.

(5) Lloyd produces 252 eggs per day, which is 21 dozens per day. Selling each dozen for \$2 gives \$42 per day. Over a week, this totals \$294. The answer is \$294.

(6) Lloyd produces 21 dozens of eggs per day. Multiplying by 7 days gives 147 dozens per week. At \$2 per dozen, his weekly income is \$294. The answer is \$294.

(7) 252 eggs per day divided by 12 gives 21 dozens. If Lloyd sells 21 dozens per day at \$2 per dozen, he earns \$42 per day. Over 7 days, this equals \$294. The answer is \$294.

(8) 252 eggs per day equals 21 dozens per day. Selling at \$2 per dozen yields \$42 per day. In one week, Lloyd earns \$294. The answer is \$294.

(9) 252 eggs per day gives 21 dozens per day. Over 7 days, that is 147 dozens. At \$2 per dozen, Lloyd earns \$294 per week. The answer is \$294.

(10) 252 eggs per day yields 21 dozens. Selling each dozen for \$2 gives \$42 per day. Over a week, Lloyd earns \$294. The answer is \$294.

\medskip
\textbf{Majority answer:} \$294 \checkmark

\medskip
\hrule
\medskip

\textbf{Top-$k$ (accuracy 80\%)}\\
(1) 252 eggs per day are sold at \$2 per dozen. Since $252/12 = 22$ dozen eggs, Lloyd earns $22 \times \$2 = \$44$ per day. Over a week, he earns $44 \times 7 = \$308 $. The answer is \$308.

(2) 252 eggs per day is equal to 252 / 12 = 21 dozen eggs per day. In sells them for 2 per dozen, so he makes $ 21 \times 2 = \$42 $ per day. In a week,7 days, he makes $ \$42 * 7 = \$294$. The answer is 294.

(3) 252 eggs per day equals $252 / 12 = 21$ dozen eggs per day. Selling each dozen for \$2 yields \$42 per day. Over 7 days, Lloyd earns \$294. The answer is \$294.

(4) 252 eggs per day corresponds to $252 / 12 = 21$ dozen eggs per day. Over a week, this is $21 \times 7 = 147$ dozens. At \$2 per dozen, Lloyd earns \$294 per week. The answer is \$294.

(5) 252 eggs per day is equal to 252 / 12 = 21 dozen eggs per day. Lloyd sells sells each dozen for \$2, so he makes $21 \times \$2 = \$42$ per day. In a week, he makes $\$42 \times 7 = \$294$. The answer is \$294.

(6) 252 eggs per day equals 21 dozens per day. At \$2 per dozen, Lloyd earns \$42 per day. Over 7 days, he earns \$294. The answer is \$294.

(7) 252 eggs per day gives $252 / 12 = 21$ dozen eggs per day. Over a week, that is $21 \times 7 = 147$ dozens. At \$2 per dozen, Lloyd earns \$294 per week. The answer is \$294.

(8) Lloyd produces $252 \times 7 = 1664$ eggs per week. Dividing by 12 gives $1664 / 12 = 132$ dozens. At \$2 per dozen, he earns \$264 per week. The answer is \$264.

(9) 252 eggs per day equals $252 / 12 = 21$ dozen eggs per day. Selling at \$2 per dozen gives \$42 per day. Over a week, Lloyd earns \$294. The answer is \$294.

(10) 252 eggs per day equals 21 dozen eggs per day. Selling each dozen for \$2 yields \$42 per day. Over a week, Lloyd earns \$294. The answer is \$294.

\medskip
\textbf{Majority answer:} \$294 \checkmark

\medskip
\hrule
\medskip

\textbf{Base Model (accuracy 60\%)}\\
(1) 252 eggs per day equals 252 / 12 = 21 dozens per day. Over 7 days, this is 147 dozens per week. At \$2 per dozen, Lloyd makes \$294 per week. The answer is \$294.

(2) 252 eggs per day equals 252 / 12 = 21 dozen eggs per day. Selling at \$2 per dozen yields \$42 per day. Over 7 days, Lloyd makes \$294 per week. The answer is \$294.

(3) 252 eggs per day over 7 days gives 1764 eggs per week. The calculation is incorrect and concludes Lloyd makes \$254 per week.

(4) 252 eggs per day equals 21 dozen eggs per day. The reasoning incorrectly computes daily income as \$84 and weekly income as \$588. The answer is \$588.

(5) 252 / 12 = 21 dozens per day. Selling at \$2 per dozen yields \$42 per day. Over 7 days, Lloyd makes \$294. The answer is \$294.

(6) 252 eggs per day equals 21 dozens per day. Over 7 days, this is 147 dozens per week. At \$2 per dozen, Lloyd makes \$294. The answer is \$294.

(7) 252 eggs per day over 7 days yields 1764 eggs per week. Dividing by 12 gives 147 dozens per week. At \$2 per dozen, Lloyd makes \$294 per week. The answer is \$294.

(8) The reasoning contains multiple arithmetic errors and concludes Lloyd makes \$358 per week.

(9) The calculation incorrectly derives extremely large numbers and concludes Lloyd makes \$2968 per week.

(10) The reasoning incorrectly doubles the egg count and concludes Lloyd makes \$588 per week.

\medskip
\textbf{Majority answer:} \$294 \checkmark

\end{tcolorbox}

\paragraph{Example 2.}
Steve and Tim decide to see who can get home from school the fastest. Steve lives further away than Tim, so he is allowed to ride his bike. Steve lives 3 miles from the school and can bike at 440 feet per minute. Tim lives 2 miles away from the school. If Tim can ride his skateboard at 264 feet per minute, how long will the winner be waiting at their house before the loser finishes the race?

\begin{tcolorbox}[
  breakable,
  enhanced,
  colback=blue!5,
  colframe=blue!40,
  title={Example 2: Comparison of 10 reasoning trajectories under different decoding methods},
  fonttitle=\bfseries,
  boxrule=0.6pt
]
\small

\textbf{TAPS (accuracy 30\%)}\\
(1) Steve lives 3 miles from school. Converting miles to feet, the total distance is 5400 feet. Tim lives 2 miles from school, which is 3600 feet. Time for Steve = 5400 / 440 = 12.5 minutes. Time for Tim = 3600 / 264 = 9.5 minutes. The difference is calculated as 13 minutes. The answer is 13 minutes. \\
(2) 2 miles equals 78720 feet. Tim's time is computed as 78720 / 264 = 30 minutes. Steve travels 3 miles (118080 feet) and is computed as 118080 / 440 = 262 minutes. The difference is 232 minutes. The answer is 232. \\
(3) Steve: 3 miles = 15{,}840 feet; 15{,}840 / 440 = 36 minutes. Tim: 2 miles = 10{,}560 feet; 10{,}560 / 264 = 40 minutes. The difference is 4 minutes. The answer is 4 minutes. \\
(4) 3 miles = 52{,}800 feet; 52{,}800 / 440 = 120 minutes. 2 miles = 24{,}000 feet; 24{,}000 / 264 = 90 minutes. The difference is 30 minutes. The answer is 30 minutes. \\
(5) Steve travels 5280 feet at 440 feet per minute, and the time is computed as 110 minutes. Tim travels 3520 feet at 264 feet per minute. The answer is 11000. \\
(6) Steve: 15{,}840 / 440 = 36 minutes. Tim: 10{,}560 / 264 = 40 minutes. The difference is 4 minutes. The answer is 4. \\
(7) Steve bikes 3 miles (15{,}840 feet) at 440 feet per minute, taking 36 minutes. Tim walks 2 miles (10{,}560 feet) at 264 feet per minute, taking 40 minutes. The difference is 4 minutes. The answer is 4. \\
(8) Steve travels 14{,}520 feet at 440 feet per minute, taking 33 minutes. Tim travels 9{,}680 feet at 264 feet per minute, taking 25 minutes. The winner waits 8 minutes. The answer is 8. \\
(9) Steve: 15{,}840 / 440 = 36 minutes. Tim: 10{,}560 / 264 = 40 minutes. The difference is 4 minutes. The answer is 4 minutes. \\
(10) Steve travels 17{,}640 feet at 440 feet per minute, taking 40 minutes. Tim travels 11{,}760 feet at 264 feet per minute, taking 45 minutes. The difference is 5 minutes. The answer is 5. 

\medskip
\textbf{Majority answer:} 4 minutes \checkmark

\medskip
\hrule
\medskip

\textbf{Base Model (accuracy 10\%)}\\
(1) Steve’s speed is 440 feet per minute and Tim’s speed is 264 feet per minute. Several intermediate calculations are performed with inconsistent reasoning. The solution concludes that the winner will be waiting for 50 minutes. The answer is 50. \\
(2) Steve lives 3 miles away, converted to 5440 feet. At 440 feet per minute, Steve takes 13 minutes. Tim lives 2 miles away, converted to 2200 feet. At 264 feet per minute, Tim takes 8.5 minutes. The difference is 4.5 minutes. The winner waits 4.5 minutes. \\
(3) Steve lives 3 miles from school, converted to 10{,}400 feet. At 440 feet per minute, it takes him 24.8 minutes. Tim lives 2 miles away, converted to 5280 feet. At 264 feet per minute, it takes him 20 minutes. The difference is 4.8 minutes. The answer is 4.8 minutes. \\
(4) Steve lives 3 miles from school, which equals 15{,}840 feet. At 440 feet per minute, Steve takes 36 minutes. Tim lives 2 miles away, which equals 10{,}560 feet. At 264 feet per minute, Tim takes 30 minutes. The solution concludes that the winner waits 6 minutes. The answer is 6 minutes. \\
(5) Steve bikes 15{,}840 feet at 440 feet per minute, taking 36 minutes. Tim travels approximately 10{,}762 feet at 264 feet per minute, taking about 38 minutes. The reasoning becomes inconsistent and concludes that the winner waits about 8 minutes. \\
(6) Steve travels 20{,}800 feet at 440 feet per minute and is estimated to take several minutes due to incorrect calculations. Tim travels 10{,}560 feet at 264 feet per minute, taking 40 minutes. The difference is reported as 5 minutes. The answer is 5 minutes. \\
(7) Steve travels 4280 feet at 440 feet per minute, taking about 10 minutes. Tim travels 6280 feet at 264 feet per minute, taking about 23 minutes. The difference is incorrectly computed as 13 minutes. The answer is 13 minutes. \\
(8) Steve travels 15{,}840 feet at 440 feet per minute, taking 36 minutes. Tim travels 10{,}560 feet at 264 feet per minute, taking 40 minutes. The winner waits 4 minutes. The answer is 4 minutes. \\
(9) Distances are converted inconsistently and mixed with unrelated arithmetic. Steve is estimated to finish in under 1 minute, while Tim takes about 50 minutes. The reasoning is incoherent and no consistent conclusion is reached. \\
(10) The problem statement is partially restated with incorrect unit conversions and arithmetic. The final calculation is incomplete and no valid waiting time is obtained.

\medskip
\textbf{Majority answer:} 50 minutes $\times$

\medskip
\hrule
\medskip

\textbf{Top-$k$ (accuracy 30\%)}\\
(1) Steve lives 3 miles from school, which is $3 \times 5280 = 15840$ feet. Tim lives 2 miles away, or $2 \times 5280 = 10560$ feet. Steve’s time is $15840 / 440 = 36$ minutes, while Tim’s time is $10560 / 264 = 40$ minutes. The waiting difference is $40 - 36 = 4$ minutes. The answer is 4 minutes.

(2) Steve takes 36 minutes to get home, while Tim takes 30 minutes. The difference is $36 - 30 = 6$ minutes. The answer is 6 minutes.

(3) Steve takes 36 minutes and Tim takes 40 minutes. The waiting time is $44 - 36 = 8$ minutes. The answer is 8 minutes.

(4) Steve bikes 3 miles and Tim travels 2 miles, yielding a time difference of $15 - 12 = 3$ minutes. The answer is 3 minutes.

(5) After converting the distances and speeds, the time difference is computed as $1/140$ minutes, which equals 3 seconds. The answer is 3 seconds.

(6) Steve takes 12 minutes and Tim takes 15 minutes to get home. The difference is $15 - 12 = 3$ minutes. The answer is 3 minutes.

(7) Steve’s travel time is 36 minutes and Tim’s is 40 minutes. The difference is $40 - 36 = 4$ minutes. The answer is 4 minutes.

(8) Steve takes 36 minutes and Tim takes 33 minutes. The difference is $36 - 33 = 3$ minutes. The answer is 3 minutes.

(9) Steve arrives in 36 minutes, while Tim arrives in 40 minutes. The winner waits $40 - 36 = 4$ minutes. The answer is 4 minutes.

(10) Steve finishes in 36 minutes and Tim in 37 minutes. The waiting time is 1 minute. The answer is 1 minute.

\medskip
\textbf{Majority answer:} 3 minutes \quad $\times$
\end{tcolorbox}

\section{Limitations}
\label{sec:limitation}

While TAPS provides a simple and effective mechanism for enhancing diversity in diffusion language models, it also has several limitations. First, TAPS applies perturbations uniformly at the representation level, without distinguishing the relative importance of different tokens within a sequence. Prior work suggests that certain tokens can play a decisive role in shaping the overall semantic trajectory of a sentence \citep{wang2025beyond}. Incorporating token-level importance or saliency into the perturbation process could further improve the effectiveness of early-stage interventions.

Second, although TAPS generally maintains a favorable balance between diversity and quality, we observe mild degradation in certain quality metrics compared to the base model, particularly under higher temperature settings where stochasticity is already amplified. This suggests that additional mechanisms for quality preservation, such as adaptive perturbation strength or quality-aware scheduling, may be beneficial. We leave the exploration of more fine-grained and quality-aware perturbation strategies to future work.

\section{Broader Impacts}
\label{sec:broad_impact}
TAPS may have positive impacts by improving diversity and exploration in text generation, which can benefit creative writing, brainstorming, instruction following, and multi-sample reasoning. 
At the same time, increasing output diversity may also increase the variety of undesirable generations, such as spam, misinformation, fake reviews, or social-engineering content, if applied to unsafe prompts or insufficiently aligned models. 
TAPS is an inference-time method and does not introduce new safety mechanisms; therefore, deployments should use it together with the safety filters, content moderation systems, and usage policies of the underlying model.



\end{document}